# Learning classifier systems with memory condition to solve non-Markov problems

Zhaoxiang Zang

The author for correspondence

*Institute for Pattern Recognition & Artificial Intelligence, Huazhong University of Science and Technology, Wuhan Hubei, 430074, China*

zxzang@gmail.com

Dehua Li

*Institute for Pattern Recognition & Artificial Intelligence, Huazhong University of Science and Technology, Wuhan Hubei, 430074, China*

jingyujingyu@yeah.net

Junying Wang

*College of Computer and Information Technology, China Three Gorges University, Yichang Hubei, 443000, China*

wannjy@163.com

**Abstract** In the family of Learning Classifier Systems, the classifier system XCS has been successfully used for many applications. However, the standard XCS has no memory mechanism and can only learn optimal policy in Markov environments, where the optimal action is determined solely by the state of current sensory input. In practice, most environments are partially observable environments on agent's sensation, which are also known as non-Markov environments. Within these environments, XCS either fails, or only develops a suboptimal policy, since it has no memory. In this work, we develop a new classifier system based on XCS to tackle this problem. It adds an internal message list to XCS as the memory list to record input sensation history, and extends a small number of classifiers with memory conditions. The classifier's memory condition, as a foothold to disambiguate non-Markov states, is used to sense a specified element in the memory list. Besides, a detection method is employed to recognize non-Markov states in environments, to avoid these states controlling over classifiers' memory conditions. Furthermore, four sets of different complex maze environments have been tested by the proposed method. Experimental results show that our system is one of the best techniques to solve partially observable environments, compared with some well-known classifier systems proposed for these environments.

## 1. Introduction

Learning Classifier System (LCS) is a rule-based machine learning technique based on Genetic Algorithm and reinforcement learning. It can map the input stimuli to output actions by a population of compact and general "condition-action-payoff rules", called classifiers. Through an evolutionary mechanism, LCS can also learn to adapt to new or changing environments. The original Learning

Classifier System framework(Goldberg 1989; Holland 1975; Holland and Reitman 1977; Liepins et al. 1991; Wilson and Goldberg 1989) proposed by Holland, is referred to as the traditional framework now. While the classifier system XCS is a special LCS proposed by Wilson(Wilson 1995) that differs mainly from Holland's framework in that classifier fitness is based on the accuracy of the prediction instead of the prediction itself. The classifier system XCS has solved the former main shortcoming of LCSs, which is the problem of strong over-generals(Kovacs 2000), by its accuracy based fitness approach.

However, according to Wilson's original proposal, XCS is not like Holland's framework, as it does not include an internal message list and no other memory mechanism either. Thus, XCS can only learn optimal policy in Markov environments, where the optimal action depends only on the state of current sensory input, and the agent can determine the states of the environment completely. But in many applications, the agent has only partial information about the current state of the environment. So the agent sensors cannot completely determine the exact states of the environment, and some different environmental states will be perceived as the same one by the agent's sensation. The agent is then said to suffer from the hidden state problem or perceptual aliasing problem, while the environment is said to be partially observable or non-Markov with respect to the agent(Kaelbling et al. 1996). To deal with an environment that is partially observable, a memory mechanism is usually introduced to cope with the lack of information deriving from the sensors(Kaelbling et al. 1996), since optimal policy cannot be determined only referring to the current inputs.

Holland's original framework has an internal message list, where the system can, in principle, store information in it. In other words, we can take the message list as the temporary memory to solve problems that require internal memory. However, Holland's framework has shown only limited success on non-Markov environments(Smith 1994).

Besides, there are several attempts to apply LCSs to non-Markov problems after that, relying on different approaches to the problem, and Section 3 will give a brief review. In this paper, we will focus on some methods based on XCS, which add explicit internal states or local storage to the perceptions to cope with non-Markov problems(Landau and Sigaud 2008) . So we will expand this topic only from XCS next.

The first extension to XCS was proposed by Wilson(Wilson 1994, 1995), and implemented by Lanzi(Lanzi 1998a). Their proposal added an internal state to XCS as an internal memory register, which requires only one or a few bits; and then extended XCS's classifier with an internal condition and an internal action, which are used to sense and act on the internal register. The resulting system "XCSM" has been applied with different sizes of internal memory to non-Markov environments with two and four aliasing positions(Lanzi 1998a). His experimental results show that, XCSM converges to an optimal solution in simple environments, even if redundant bits of memory are employed. However, experiments with Maze7 (depicted in Fig. 4(b)) show that in complex problems, XCSM's exploration strategy is not enough to guarantee the convergence to an optimal solution. Thus, Lanzi devel-oped an enhanced version of XCSM, named “XCSMH”(Lanzi 1998b; Lanzi and Wilson 2000). Compared with XCSM, XCSMH consists of (i) a different policy for updating the contents of the internal memory and (ii) a hierarchical exploration strategy to select actions. Lanzi’s experiments show that XCSMH can evolve an optimal policy for most maze environments previously presented. Especially, when applied to a more difficult environment, such as Maze10 (shown in Fig. 11(b)), XCSMH still converges to a near optimal solution, outperforming XCSM(Lanzi 1998b; Lanzi and Wilson 2000).

And then, Hamzeh et al. developed two distributed-architecture classifier systems, named Parallel Specialized XCS (PSXCS)(Hamzeh and Rahmani 2008) and Recursive PSXCS (RPSXCS)(Hamzeh et al. 2009). PSXCS and RPSXCS detect aliased environmental states and assign their handling to some subordinate XCS. So, in their systems, some special sub-XCSs equipped with history windows are responsible for each detected aliasing state respectively, which causes the structure of the systems is somewhat complicated. Experimental results show that their systems can perform optimally in some well-known benchmark problems.

In order to further improve the performance of XCS within non-Markov problems, we will discuss a different way to include memory within XCS. Our proposal adds XCS an internal message list as the memory list, in which the length of a list element equals to that of strings from the detectors. Then, only extends a small number of classifiers with memory conditions, to cope with non-Markov states within the environments. Besides, in order to make the memory condition match the earlier environmental state, not just the state at the last time step, our system uses the memory list to store some recent experiences, i.e. recent perception inputs from detectors and, if necessary, corresponding actions for these perceptions. But the memory condition must keep away from those aliasing states in the memory list. Because if the memory condition of a classifier matches an aliasing state in the memory list, this classifier will be confused at the aliasing state, and have no benefit to the system[1]. Therefore, to overcome this problem, a method is needed to detect those aliasing states. We introduce and improve an aliasing state detection method(Dung et al. 2008; Hamzeh et al. 2009; Hamzeh and Rahmani 2008) to our system. With the help of the detection method, aliasing states in the environment will be recognized, thus the memory condition can always avoid these states and become efficient and effective. From the above, our proposed method is based on XCS, including an internal memory list, an extended condition in the classifier (memory condition), and an aliasing state detection mechanism. Later, we will refer to our proposal as “XCSMD”, and the “MD” stands for “memory” and “detection”. Experimental results show that XCS with memory adopting our method can perform optimally or near-optimally in partially observable environments compared with some of the best classifier systems. Additionally, XCSMD holds a clear memory mechanism, which is easy to understand.

[1] The more explanations will be given in Section 4.

The rest of the paper is organized as follows. Section 2 reviews XCS. Section 3 briefly introduces the maze environments and the perceptual aliasing problem, including some related work on it. The implementation of our proposal, "XCSMD", is described in Section 4. Experiments with XCSMD are presented in Section 5 and Section 6. Finally, Section 7 ends the paper, presents our main conclusions and some directions for future research.

# 2. XCS Classifier System

XCS is a new form of Learning Classifier System that uses accuracy as a means of fitness for selection within Genetic Algorithm(Wilson 1995). Although we assume a basic familiarity with LCS and XCS, this section provides a brief outline of XCS. Further information can be found in (Lanzi 2002; Sigaud and Wilson 2007; Wilson 1995), and the algorithmic description of XCS is in (Butz and Wilson 2002).

As in other LCSs and reinforcement learning methods, XCS acts as a learning agent that perceives inputs describing the current environmental state, responds with actions, and receives a reward (from a separated reinforcement program or from XCS itself) as an assessment of its action. However, XCS differs from Holland's classifier system in three major points(Lanzi 1998a). First, XCS's classifier fitness is based on the accuracy of the prediction, rather than the prediction itself. The traditional strength parameter of a classifier is discarded, and replaced by three parameters: (i) the prediction $p$ which gives an estimate of the payoff that the system is expected to gain when the classifier is used; (ii) the prediction error $\varepsilon$, evaluating how much precise is the prediction $p$; (iii) the fitness $F$ that evaluates the accuracy of the prediction given by $p$ and thus is a function of the prediction error $\varepsilon$. Second, compared with the original framework, XCS has no internal message list, and no other memory mechanisms. Finally, the genetic algorithm in XCS is applied to environmental niches, instead of the panmictical GA. XCS works as follows.

When XCS receives an environmental input $s_t \in \{0,1\}^L$, where $L$ is the number of bits of the detector, it forms the match set [M] of classifiers whose conditions match the input $s_t$. Match set [M] is a subset of the whole population [P] of classifiers. At least $\theta_{mna}$ actions must be present in the match set. If not, covering classifiers will be created with a matching condition. For each possible action $a \in A = \{a_1, \cdots, a_n\}$ in the match set, the system prediction[2] $PA(s_t, a)$ is computed as the fitness weighted average of the classifier predictions that advocate the action $a$ in [M].

---

[2] The variable $s_t$ is added to highlight system prediction is dependent on the current input $s_t$ since the match set [M] is determined by $s_t$. This will provide convenience for later use.

$$PA(s_t,a)=\frac{\sum_{cl.a=a\wedge cl\in[M]} cl.p\cdot cl.F}{\sum_{cl.a=a\wedge cl\in[M]} cl.F} \quad (1)$$

$cl$ stands for a classifier, $cl.p$ for payoff prediction of $cl$, and $cl.F$ for its fitness. $PA(s_t,a)$ gives an evaluation of the expected payoff if action $a$ is performed. Next, XCS selects an action from those advocated by the classifiers in [M]. Action selection can be deterministic, the action with the highest system prediction is chosen, or probabilistic, the action is chosen with a certain probability among the actions. Usually, XCS chooses actions randomly during learning and chooses the best action $a_{\max}=\arg\max_a PA(s_t,a)$ during testing. The subset of [M] which advocates the selected action is called the action set [A]. The selected action is then performed in the environment, and a scalar reward $r_t$ is returned to the system together with a new input configuration $s_{t+1}$.

In each cycle, XCS updates the classifiers in the action set $[A]_{-1}$ corresponding to the previous time step $t-1$ based on the reward received. Besides, dependent on the threshold $\theta_{ga}$ and the average time in $[A]_{-1}$ since the last GA application, a reproductive event is triggered, in which a GA is called upon to modify the population of classifiers. Since the GA in XCS only reproduces classifiers currently in $[A]_{-1}$, it realizes an implicit niching(Wilson 1998). The GA selects two classifiers with probability proportional to their fitness and copies them. It performs crossover on the copies using probability $\chi$ while using probability $\mu$ to mutate each allele. Furthermore, subsumption deletion acts in $[A]_{-1}$ deleting more specific classifiers if an accurate, experienced, and more general classifier exists. If the number of classifiers in the population [P] exceeds the threshold $N$, excess classifiers are deleted. Classifiers for deletion are selected proportionally to their action set size estimate $as$. Moreover, if sufficiently experienced (experience parameter $exp$, determines how many times the relevant classifier participates in [A]) and with a sufficiently low fitness $F$, the probability of deletion is increased further.

The method that XCS updates the classifiers in the action set $[A]_{-1}$ differs from traditional strength-based systems(Goldberg 1989; Wilson 1994), in which the fitness of a rule is called its strength, used in both action selection and reproduction. The accuracy-based XCS maintains two separate estimates of classifier's utility for action selection and reproduction. In multi-step tasks, the update process in the action set $[A]_{-1}$ works as follows.

First, the Q-learning-like payoff $P$ is computed as the sum of the reward received at the previous time step and the maximum system prediction of current time step(Wilson 1995) discounted by a factor $\gamma(0\le\gamma<1)$.

$$P_{t-1}=r_{t-1}+\gamma\max_{a\in A} PA(s_t,a) \quad (2)$$

where $r_{t-1}$ is the reward at the previous time step, $PA(s_t, a)$ is the system prediction at the current time step. Then, $P_{t-1}$ is used to update the prediction $p$ for each classifier in the action set $[A]_{-1}$ by the Widrow-Hoff delta rule(Widrow and Hoff 1988) with learning rate $\beta(0 \le \beta \le 1)$:

$$p \leftarrow p + \beta(P_{t-1} - p) \quad (3)$$

In the same way, the prediction error $\varepsilon$ is adjusted by:

$$\varepsilon \leftarrow \varepsilon + \beta(|P_{t-1} - p| - \varepsilon) \quad (4)$$

To update the fitness, two additional parameters are needed. The first one is the absolute accuracy $\kappa$ of a classifier, and it is derived from the reward prediction error $\varepsilon$ as follows:

$$\kappa = \begin{cases} 1 & \text{if } \varepsilon < \varepsilon_0 \\ \alpha(\varepsilon / \varepsilon_0)^{-\nu} & \text{otherwise.} \end{cases} \quad (5)$$

In which, the parameter $\varepsilon_0 (\varepsilon_0 > 0)$ controls the maximal tolerance for prediction error $\varepsilon$; $\alpha(0 < \alpha < 1)$ and $\nu(\nu > 0)$ are constants controlling the rate of decline in accuracy $\kappa$ when $\varepsilon_0$ is exceeded. The second parameter is the relative accuracy $\kappa'$, which comes from the absolute accuracy $\kappa$:

$$\kappa' = \kappa \Big/ \sum_{[A]_{-1}} \kappa \quad (6)$$

Finally, the fitness parameter is adjusted by the rule:

$$F \leftarrow F + \beta(\kappa' - F) \quad (7)$$

Essentially, $\kappa$ measures the current absolute accuracy of a classifier using a power function with exponent $\nu$ to further prefer low error classifiers. In other words, classifiers whose error estimate $\varepsilon$ drops below threshold $\varepsilon_0$ are considered to be accurate(Butz et al. 2001). $\alpha$ causes a strong distinction between accurate and not quite accurate classifiers.

To summarize, XCS's fitness behaves inversely to the payoff prediction error, and if the error is below $\varepsilon_0$, it will be ignored entirely. XCS also holds and updates some other parameters, which will not be described here.

# 3. The Maze Environments and the Perceptual Aliasing Problem

Two types of testing environments have been employed to study LCSs: state environments and maze environments. State environments are similar to directed graphs or state diagram. Its nodes represent the agent's sensations (i.e. environmental states) while its edges usually labeled with the action and sometimes with a transition probability, represent the effect of actions in each state. The agent should learn the shortest path to goal states. Maze problems, usually represented as grid-like two-dimensional areas that may contain different objects of any quantity and with different properties (for example, obstacle, goal, or can be empty), serve as a simplified virtual model of the real environ-ment, and can be used for developing core algorithms of many real-world applications related to the problem of navigation. However, the two types of environments are usually equivalent in most cases, except that maze environments are more intuitive and clearer than state environments(Lanzi and Wilson 2000). Besides, different complexity of mazes available from (Bagnall and Zatuchna 2005), are good testing beds for LCSs research. Therefore, in this paper, we will use mazes as the testing environments.

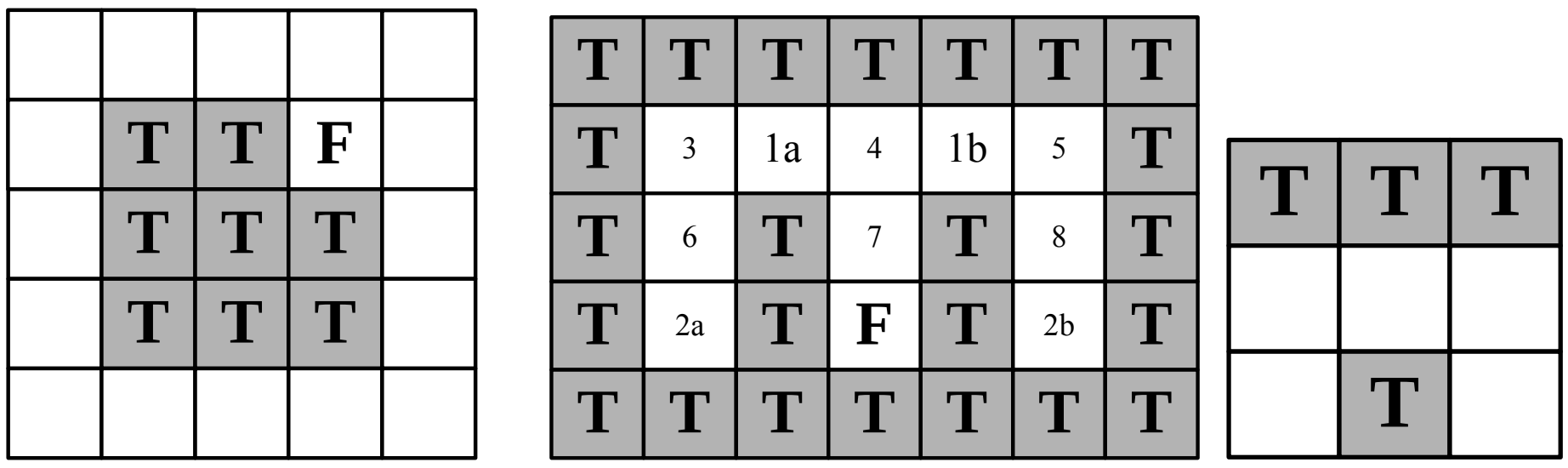

**(a)** **(b)** **(c)**

**Fig. 1** (a) Woods1 maze environment, food object is marked with F, and obstacle is marked with T. (b) Woods101 environment and its aliasing state (c). The two aliasing squares are marked with 1a and 1b.

LCSs have been the most widely used class of algorithms for reinforcement learning in mazes for the last twenty years. And it is not surprising that LCSs have presented the most promising performance results(Bagnall and Zatuchna 2005; Zatuchna and Bagnall 2009). Usually, LCSs have been applied widely to mazes where the agent can detect every square uniquely. That is to say, these mazes are Markov environments. Fig. 1(a) presents Woods1(Wilson 1994) maze environment. The maze may contain different obstacles in any quantity, such as T standing for tree in Woods1, and some objects for learning purposes, like virtual food F, which is the agents' goal to reach. It must be noted that, if a maze has not enough obstacles to mark its boundary, the left and right edges of the maze are connected, as are the top and bottom. In this paper, the agent is randomly placed in the maze on an empty cell, and the agent has two Boolean sensors for each of the eight adjacent squares. The agent can move into any adjacent square that is free. As is known from Fig. 1(a), the agent can disambiguate every square uniquely in Woods1 with respect to agent's sensation, and Woods1 is Markov environment.

However in many cases, the agent may have only partial information about the current state of the environment, so that the agent sensors cannot determine the state of the environment completely. Such environments are non-Markov and they form the most general class of environments. It is worth noting that, whether an environment is Markov or non-Markov, depends on the interaction between the environment and the agent's sensation, rather than the environment itself only.

The Woods101 maze(Cliff and Ross 1994; Mccallum 1993), depicted in Fig. 1(b), is non-Markovian since it has two distinct positions, indicated by the two same number, which the agent senses as identical but require different optimal actions. The optimal action in the right aliased position is "go south-west"; in the left aliased position it is "go south-east". When the agent is in one of the two positions, it cannot decide which action is correct solely considering its current sensation.

If the perceptual aliasing problem needs to be resolved, we should go after the most challenging and interesting mazes, which involve some squares where the limited perception of the agent makes them appear identical. LCS researchers have put some effort into disambiguating these apparently identical aliasing squares by several kinds of approaches:

- The first and common one is to add explicit internal states or local storage to the perceptions to help the decision-making process of the system, namely *Explicit Internal State Management*(Landau and Sigaud 2008). This approach was first used by Holland in the original LCS framework(Holland and Reitman 1977) with the use of an internal message list, but showed only limited success on non-Markov problems(Smith 1994). And afterward, a different form of the explicit internal state solution was adopted in ZCSM(Cliff and Ross 1994) and XCSMH(Lanzi 1998a; Lanzi and Wilson 2000; Moioli et al. 2008) by using memory register.
- The second one is the memory window or history window method, which is a special case of explicit internal state management where the internal state consists of some direct memories of the past(Landau and Sigaud 2008). The memory window can be set to fixed size(Hamzeh and Rahmani 2008), or variable size(Hamzeh et al. 2009; Mccallum 1996). In addition, our proposed system XCSMD described in Section 4, you will find, is a variant of the memory window approach with variable window size.
- The third one is to link some classifiers often fired in sequences, namely *Classifier Chaining*(Landau and Sigaud 2008), making one classifier's activation depend on the classifier previously fired, so as to use a memory of previous actions to disambiguate the current state. It links classifiers into "behavioral sequences", to form action persistence. In fact, this approach can be seen as an alternative view of the variable-sized memory window method. It is promising for solving some perceptual aliasing problems, because behavioral sequences starting before the ambiguous states and ending after these states have been crossed, can bridge aliasing states(Landau and Sigaud 2008). That is to say, the ambiguous states are hidden in sequences, and the agent never needs to guess what to do in an ambiguous context. This approach has been used in ZCCS(Tomlinson and Bull 1999) and CXCS(Tomlinson and Bull 2002), as well as Anticipatory Classifier System (ACS) that includes a mechanism to predict the next state of the environment(Métivier and Lattaud 2003; Stolzmann 1999, 2000). Recently, the Adapted Pittsburgh Classifier System (APCS)(Gilles and Mathias 2008; Gilles and Mathias 2010) used in non-Markov problems can also be considered to fall into this category.
- The fourth one is hybrids of several methods above. For example, Zatuchna and Bagnall's "AgentP" system with associative perception(Zatuchna 2005; Zatuchna and Bagnall 2009), could be considered as a hybridization of basic ACS structure with a memory mechanism, although there are significant differences in performance, reinforcement and learning mechanisms. It is based on psychological theories of the processes of perception and association in humans and animals, and better equipped to learn how to navigate in complex aliasing maze environments.

From the review above, plenty of methods and systems are available for coping with the non-Markov environments, but there are still some maze problems hard for LCS to solve in practice. What internal properties and features may hinder LCS from solving those certain types of mazes? We need a brief introduction about the analysis and classification on the complexity of maze problems to answer this question in some degree.

## 3.1 The Analysis and Classification on the Complexity of Maze Problems

First, Wilson proposes a three-classes scheme to classify reinforcement learning environments with respect to the sensory capabilities of an agent(Wilson 1991). An environment belongs to Class 0 if the agent is not only able to determine the entire state of the environment, but also can get positive reinforcement every time step as long as taking appropriate action. This might be called a pure stimulus-response environment. An environment belongs to Class 1 if the sensory capabilities of the agent are enough to determine the entire state of the environment, but the agent will suffer delayed reward. This could be called a stimulus-response environment with sparse or deferred reinforcement. However, in Class 2 environments, the agent has only partial information about the entire state of the environment. Class 2 environments are non-Markov with respect to agent's actions, and will suffer delayed rewards.

Littman presents a more formal classification of reinforcement learning environments, based on the simplest agent that can achieve optimal performance(Littman 1993). Two parameters $h$ and $\beta$ characterize the complexity of an agent. An $(h,\beta)$ environment is best solved by an $(h,\beta)$ agent that uses the input information provided by the environment and at most $h$ bits of local storage to choose the action which maximizes the next $\beta$ reinforcements. Therefore, Class 0 environments correspond to $(h=0;\beta=1)$ environments and Class 1 to $(h=0;\beta>1)$ environments, while Class 2 environments correspond to $(h>0;\beta>1)$ (non-Markov) environments.

Recently, Zatuchna and Bagnall have done much analysis on the complexity of maze problems(Bagnall and Zatuchna 2005), especially on Class 2 environments. And they have introduced new metrics for classifying the complexity of mazes. Next, there is an introduction about their metric for classifying the types of aliasing and the influence of aliasing type on maze complexity.

They use two major factors, which have a significant influence on the learning process, to classify the types of aliasing. The first factor is minimal distance to food, $d$ ; the second is correct direction to food, or appropriate action, $a$ . Let $d_1$ and $d_2$ be minimal distance to food from an aliasing square 1 and aliasing square 2 respectively, and $a_1$ and $a_2$ be the optimal actions for the two squares respectively. Square 1 and square 2 belong to a same aliasing state. They introduce four different situations for division:

- Both distance and action are the same $(d_1 = d_2, a_1 = a_2)$; the squares are pseudo-aliasing.
- Distance is different but action is the same $(d_1 \neq d_2, a_1 = a_2)$; type I aliasing squares.
- Distance is different and action is different $(d_1 \neq d_2, a_1 \neq a_2)$; type II aliasing squares.
- Distance is the same but action is different $(d_1 = d_2, a_1 \neq a_2)$; type III aliasing squares.

In Fig. 1(b), the two pseudo-aliasing squares of Wood101 is marked with 2a and 2b. Seen from Fig. 2, the aliasing squares of Littman57(Littman et al. 1995) marked with 1, 2 and 3 have the same directions to food respectively (aliasing type I). Fig. 3(b) shows MazeF4(Stolzmann 2000) with type II aliasing squares marked with 1, which have different distances to food as well as different directions. Woods101 in Fig. 1 is an example of mazes with type III aliasing squares.

| | | | | | | | | | | | | |
|---|---|---|---|---|---|---|---|---|---|---|---|---|
| **T** | **T** | **T** | **T** | **T** | **T** | **T** | **T** | **T** | **T** | **T** | **T** | **T** |
| **T** | 4 | 5 | 2a | 3a | 2b | 3b | 2c | 6 | 7 | 8 | 9 | **T** |
| **T** | **T** | **T** | 1a | **T** | 1b | **T** | 1c | **T** | **F** | **T** | **T** | **T** |
| **T** | **T** | **T** | **T** | **T** | **T** | **T** | **T** | **T** | **T** | **T** | **T** | **T** |

**Fig. 2** Littman57 environment, type I aliasing maze. The three aliasing states are marked with 1, 2, and 3. The other six non-aliasing states are marked with 4 to 9.

For maze that has combined aliasing (more than one aliasing type), Zatuchna and Bagnall define its type by the highest aliasing type the maze contains. Then, they have analyzed a big mazes collection used in the literature, resulting that for most learning agents, type III aliasing mazes may be considered as the most difficult group of aliasing mazes, type II mazes are of medium complexity and those type I are the easiest(Bagnall and Zatuchna 2005). In addition, aliasing conglomerate, which is a group of adjacent aliasing squares, i.e. accumulation of several aliasing squares side by side, will cause particular difficulty for learning agents. If some aliasing conglomerates of a similar graphic pattern locate in different areas of a maze, which Zatuchna refers to as aliasing clones(Bagnall and Zatuchna 2005; Zatuchna and Bagnall 2009), it will be an even more complex phenomenon. For example, as is shown in Fig. 2, there are two aliasing clones involving 1-2-3 and two similar ones involving 3-2-1 states in Littman57; Woods102(Cliff and Ross 1994) in Fig. 4(a) has two chains that consist of adjacent aliasing squares 1 and 2. Other mazes, as Woods101 (Fig. 1), may have communicating aliasing squares, that is two aliasing cells bordering on the same (non-aliasing) neighbor square. The existence of aliasing conglomerates or communicating aliasing squares increases the complexity of the learning task significantly.

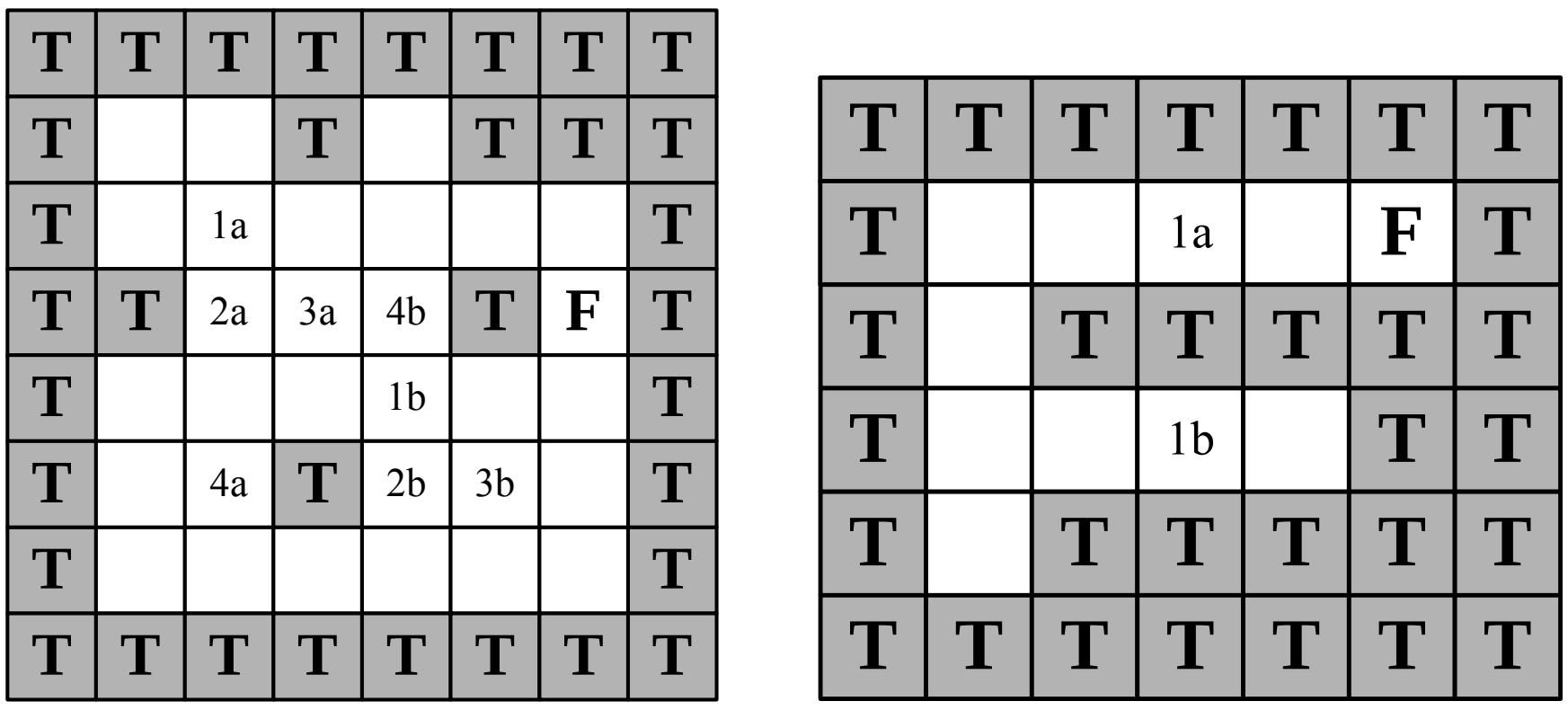

**(a)** **(b)**

**Fig. 3** (a) MiyazakiA environment, type I aliasing maze; (b) MazeF4 environment, type II aliasing maze.

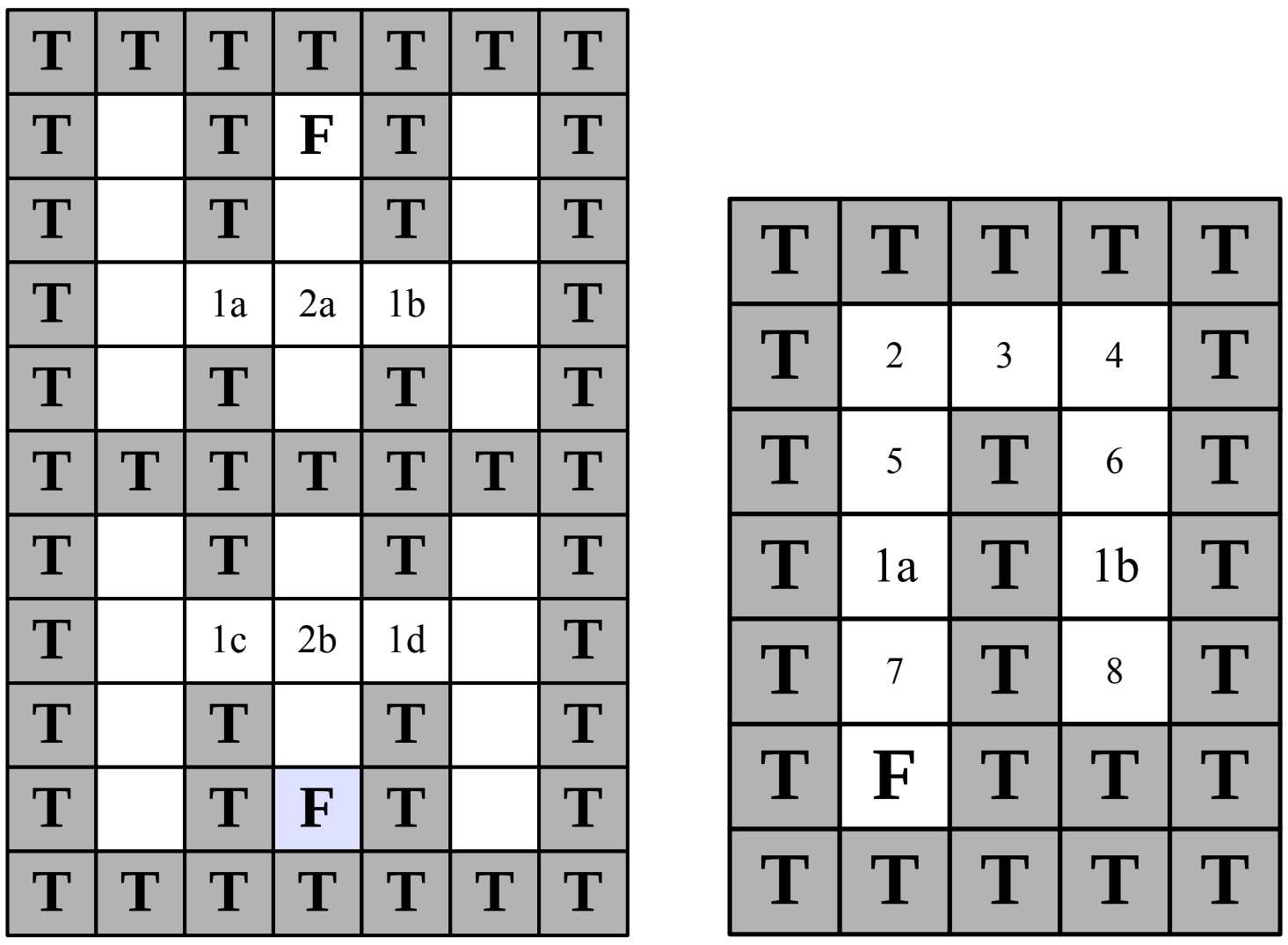

**(a)** **(b)**

**Fig. 4** (a) Woods102 environment, aliasing type III; (b) Maze7 environment, aliasing type II.

# 4. Adding Memory Mechanism to XCS

As described in Section 1, our proposal system, XCSMD, mainly consists of (i) adding an internal message list to XCS as the internal memory list, which is used to store some earlier environmental states, (ii) extending classifier with a memory condition, which is used to sense a specified element in the memory list, and (iii) introducing an aliasing state detection method to XCS. Next, the detailed description of its structure will be given.

## 4.1 The Structure of the Classifiers

XCSMD's classifier consists of three parts:

- A normal condition part, or current condition part $c \in \{0,1,\#\}^L$, with attributes of environmental states that define some sensory situations, is used to match the strings from the detectors at current time step, as the external condition in Lanzi's XCSMH.
- A memory condition part $m \in \{0,1,\#\}^L$, with attributes of environmental states that define some previous sensory situations and, if necessary, corresponding action responded by the agent. It is used to match a specified element in the memory list, as the internal condition in XCSMH.
- An action part $a \in A = \{a_1, \cdots, a_n\}$, with instructions for the effectors, as the external action in XCSMH, but not including an internal action.

Thus, a $(m,c,a)$ -classifier can be interpreted as follows:

If the current environmental state has the attributes of $c$, and a specified element in the memory list has the attributes of $m$, then the action $a$ may be sent to the effectors.

However, we notice that not every state in a non-Markov environment is an aliasing state, and usually the number of aliasing states is in the minority compared with the total number of states. There is no need to make every classifier contain a memory condition. So, we will introduce two types of classifiers to XCSMD:

<memory condition><normal condition>:<action>

<normal condition>:<action>

One type consists of memory condition, normal condition and action. The other only consists of normal condition and action. The classifiers with memory conditions are prepared for aliasing states, while the normal classifiers are mainly for the other states.

In order to distinguish the two types of classifiers, an additional integer parameter, called memory pointer $mp$, must be added to every classifier. If $cl.mp$ [3] equals to -1, it indicates that $cl$ is a classifier without memory condition; if $cl.mp$ is greater than -1, $cl$ is a classifier with memory condition. Besides, when $cl.mp$ is greater than -1, the value of $cl.mp$ also indicates which element in the memory list will be matched by the memory condition of $cl$.

The memory condition is actually a history window, whose size is fixed and just set to one, and can only look back (recall) one specified state in the past, but not just the state at the last time step. So, it can also be seen as a history window with variable window size.

---

[3] To refer to one of the attributes of a classifier $cl$, we use the dot notation. For example, $cl.mp$ is used to refer to $mp$ of $cl$.

## 4.2 Internal Memory List

In order to make the memory condition at current time step $t$ match the earlier environmental message, not just the environmental message $s(t-1)$ at the last time step, XCSMD uses the memory list $ml$ to store some past environmental messages. Precisely, it is to record some recent experiences of the agent, i.e. recent perception inputs $s$ from detectors and, if required, corresponding actions $a \in A$ for these perceptions. The size of the memory list $ml$ is defined as $N_{ml}$ $(N_{ml} > 0)$. That is to say, the number of the past environmental messages that the memory list can store is $N_{ml}$. The memory list is actually a queue data structure in XCSMD. The environmental message enters at the rear of the list and is removed from the front of the list. The two primary operations involving the memory list are adding a new environmental message to the list and removing a message from the list, which must follow the first-in-first-out rule. The environmental messages are stored in the order in which they occur, and every element in the memory list has an index $i$ to indicate its order in the list, so we can use $ml[i]$ to refer to the $i^{th}$ element. For example, $ml[0]$ may stand for the environmental message $s(t-1)$ at the previous time step, and $ml[1]$ may stand for $s(t-2)$.

The update process of the internal memory list $ml$ works as follows. At the end of the current time step $t$, $ml$ will be updated by the environmental message $s(t)$ detected at current time step, that is the message $s(t)$ will be added to $ml$. If the number of messages in $ml$ exceeds the threshold $N_{ml}$, excess element is deleted. At the beginning of a problem (trial), the internal memory list $ml$ will be set to empty. It must be noted that the update of the memory list should be controlled strictly. If the environmental message $s(t)$ in the current time step is the same as the one $s(t+1)$ detected at the next time step, the message $s(t)$ is not required to update $ml$, but leave it unmodified, since that no change between the environmental message $s(t+1)$ and $s(t)$ always suggests that the agent does not move actually at the current time step $t$. Therefore, the index $i$ of the memory list element $ml[i]$ does not correspond completely to the time step at which the element is added to the memory list.

From the above, it is clear that the length of the normal condition part $c$ is equal to that of strings from the detectors. The memory list element $ml(i)$ is the same length as the memory condition part $m$, and its length equals to the total length of environmental message and the action part $a$ ( if $a$ is included in $m$ to as part of the memory). The relationship between them can be summarized as Equation (8) and (9).

$$strlen(c) = strlen(s(t)) = L \tag{8}$$

$$strlen(ml(i)) = strlen(m) = \begin{cases} L + strlen(a) & \text{if } a \text{ is included in } m, \\ L & \text{otherwise.} \end{cases} \tag{9}$$

## 4.3 Aliasing State Detection

As described in Section 1, when the agent faces an aliasing state, the memory condition is required to match a former environmental state in the memory list to disambiguate the aliasing state. But the memory condition must keep away from those aliasing states in the memory list. If the memory condition of a classifier matches an aliasing state in the memory list, the memory condition as a foothold will fail to determine the exact location of the agent, and cannot decide which direction the agent should go along, since it will be confused at the aliasing state. In such a case, the memory condition will have no benefit to the system.

Therefore, a method is needed to detect those aliasing states. We introduce and improve an aliasing state detection method to XCSMD from (Dung et al. 2008), (Hamzeh and Rahmani 2008) and (Hamzeh et al. 2009). With the help of this method, the aliasing states in the environment will be recognized, and stored in the aliasing state list $asl$. This list holds all the aliasing states detected by this approach, and will be looked up by other processes to determine whether a state is aliased. Next, we will describe the aliasing state detection method used in XCSMD, which is different from the original method in some respects. Some more detailed description and explanation of this method can be found in the three papers.

In an aliasing state, there are at least two optimal actions to choose from but none of them give stable feedback (reward). In other words, the stable feedback should be low for an aliasing state, and should be high for a fully observable state (non-aliasing state). So we will take the stable feedback value $\nu(s,a)$ as the criterion to detect Aliasing State Relevant (ASR) classifiers, which indicate those aliasing states by their condition part. But first we need to use classifier's specificity $spe$ to distinguish ASR classifier from over-general classifier, since an ASR classifier only corresponds to an environmental state and can only vote an aliasing state. The specificity $spe$ of a classifier $cl$ is defined as:

$$cl.spe = \frac{\text{number of non-\# positions in } cl}{\text{length of classifier } cl} \tag{10}$$

The rule to detect ASR classifiers is: if a classifier's specificity is equal to '1', it is sufficiently experienced, and its stable feedback value is far from '1', then it will be an ASR classifier. We can formalize it as:

$$cl.spe = 1 \wedge cl.exp > \theta_{ASR} \wedge cl.\nu < \tau \Rightarrow cl \text{ is an ASR classifier} \tag{11}$$

$cl$ stands for a classifier, $cl.exp$ is the experience parameter of $cl$, and $cl.\nu$ is the stable feedback value $\nu(s,a)$ for $cl$. $\theta_{ASR}$ and $\tau$ are two predefined thresholds, set at 30 and 0.4 respectively in our experiments.

The stable feedback value can be estimated from the critical error. First, the critical error $\delta(s,a)$ is calculated as follow:

$$\delta(s_{t-1},a_{t-1}) = r_{t-1} + \gamma \max_{a\in A} PA(s_t,a) - \max_{a\in A} PA(s_{t-1},a) \tag{12}$$

If the critical error is positive, update the positive feedback value by

$$\delta^+(s_{t-1},a_{t-1}) = \delta^+(s_{t-1},a_{t-1}) + \delta(s_{t-1},a_{t-1}), \tag{13}$$

whereas if the critical error is negative, update the negative feedback by

$$\delta^-(s_{t-1},a_{t-1}) = \delta^-(s_{t-1},a_{t-1}) - \delta(s_{t-1},a_{t-1}) \tag{14}$$

Then, the stable feedback value for state $s_{t-1}$ and also for the ASR classifier is defined by

$$\nu(s_{t-1},a_{t-1}) = \frac{\left|\delta^+(s_{t-1},a_{t-1}) - \delta^-(s_{t-1},a_{t-1})\right|}{\delta^+(s_{t-1},a_{t-1}) + \delta^-(s_{t-1},a_{t-1})} \tag{15}$$

in which the stable feedback value $\nu(s,a)$, the positive feedback $\delta^+(s,a)$, and the negative feedback $\delta^-(s,a)$ are all new parameters within the ASR classifiers.

When XCSMD updates the classifiers in the action set $[A]_{-1}$, XCSMD applies the ASR classifier detection rule to recognize possible aliasing states in the environment via scanning the classifiers in the action set. During this phase, both two types of classifiers in XCSMD are responsible for the detection of aliasing state. If a classifier without memory condition satisfies the above rule, then it is an ASR classifier, and its relevant state will be added to the aliasing state list $asl$; If a classifier with memory condition satisfies the above rule, and the state corresponding to its normal condition is in the aliasing state list, then it is an ASR classifier and the state corresponding to its memory condition will be added to the list.

Every element in $asl$ has a parameter called numerosity $num$ to indicate how many times the element has been recognized as an aliasing state. When a new aliasing state needs to be inserted in the list $asl$, it is compared with existing ones to check whether there already exists a same element in the list. If such an element exists, then the new state element is not inserted in the list. Instead, the parameter $num$ of the existing element is increased. If there is no such an element in the list, then the new aliasing state is added to the list. In practice, we will take the parameter $num$ as a vote on the certainty of the relevant state element in $asl$, since some non-aliasing states may be added to the list mistakenly. If the $num$ of a state element in the list is greater than others, we will prefer to believe this state element is really an aliasing state. Here, some simple rules are used to decide whether a state element in the list $asl$ is an aliasing state or not. First, the middle value of the maximum numerosity and minimum numerosity is computed as

$$num_{\frac{1}{2}} = (num_{\max} + num_{\min})/2 \tag{16}$$

where $num_{\max}$ and $num_{\min}$ are the maximum and minimum $num$ within elements of the list $asl$. Then, the middle value $num_{\frac{1}{2}}$ is taken as the criterion to decide whether a state $s$ is an aliasing state: if $s \in asl \wedge s.num > num_{\frac{1}{2}}$, $s$ is an aliasing state; if $s \in asl \wedge s.num \le num_{\frac{1}{2}} \vee s \notin asl$, $s$ is not an aliasing state.

Alternatively, a probabilistic method can be used to determine the probability that a state $s$ is an aliasing state:

$$\Pr(s) = \begin{cases} 1 & \text{if } s \in asl \ \wedge \ s.num \geq num_{\frac{1}{2}} \\ s.num / num_{\frac{1}{2}} & \text{if } s \in asl \ \wedge \ s.num < num_{\frac{1}{2}} \\ 0 & \text{if } s \notin asl \end{cases} \tag{17}$$

When we decide whether a state $s$ is aliased, a random number $rand(0,1)$ between 0 and 1 is compared with $\Pr(s)$. If $rand(0,1) < \Pr(s)$, $s$ is an aliasing state; otherwise not. Here, this probabilistic method is adopted by us.

Next, the most important is to provide a mechanism in XCSMD to let ASR candidate classifiers be created in the population, because the specificity of ASR candidate classifiers must be equal to 1, and the system cannot develop enough ASR candidate classifiers by itself. Even worse is that the probability of automatically creating an ASR candidate classifier decrease drastically when the classifier length becomes longer. Thus, it is necessary to guarantee the existence of enough needed ASR candidate classifiers during the whole process of detection. The applying of "action set covering operator" is a solution to this problem. This operator is triggered after forming all action sets in each trial. In action set [A], if there is no classifier whose accuracy parameter $\kappa$ is equal to 1, and experience parameter $exp$ is greater than $\theta_{ASCover}$, then a classifier with the same condition as the environmental state (without any '#' symbol; specificity parameter $spe$ =1), and with the same action as the proposed action by [A] is created and inserted into the population [P]. $\theta_{ASCover}$ is a predefined parameter. It must be noted that if the current environmental state is in the aliasing state list $asl$, and the memory list $ml$ is not empty, then a classifier with memory condition will be created, otherwise a classifier without memory condition will be created. The method of creating the new classifier is the same as the covering method, which will be described in section 4.4, and the only difference is that the specificity of the new classifier equals to 1. This routine will provide enough ASR candidate classifiers for the detection mechanism. In addition, because ASR classifiers are low fitness classifiers and they may be deleted due to the nature of XCS, fitness calculation routine can be changed to make these classifiers survive in the population and discovered by the detection mechanism. To do so, if the specificity parameter $spe$ of a classifier is equal to '1' and its fitness $F$ is below the mean fitness of the whole population [P], then its fitness will be set to the mean fitness of the population.

The methods above can guarantee the existence of needed ASR candidate classifiers in the population. But this may cause a serious problem. Because the specificity parameter $spe$ of ASR candidate classifier is equal to '1', while the nature of XCS is to evolve the most accurate and general classifiers whose specificity parameters are lower than '1'. Some of these ASR candidate classifiers may not be relevant to aliasing states, and presence of them in the population has no benefit for the generalization mechanism of XCS. To cope with this problem, we must remove those unwanted classifiers with specificity equal to '1' from the population. First, if an ASR candidate classifier has been recog-nized as a true ASR classifier by the detection mechanism and its relevant environmental state has been added to the aliasing state list, then this classifier will be deleted directly from [P]. Second, the other ASR candidate classifiers which are just non-ASR classifiers will be eliminated from [P] by the subsumption deletion routine, or will be converted into some general classifiers with specificity lower than '1' by the generalization mechanism, since they are not relevant to aliasing states and don't do much harm to the system. The subsumption deletion routine includes GA subsumption and action set subsumption, which can be enabled selectively since they are not necessary.

At last, it should be noted that there is no need for the action set covering routine and the aliasing state detection routine to be running during the entire process of an experiment. These routines can be enabled only in the early stages of the experiment.

## 4.4 Covering and Matching Process

In the formation of the initial population of classifiers in XCSMD, memory condition part $M$ can be initialized by the covering operator, as the normal condition part. There are two types of classifiers in XCSMD, thus covering procedure is required to be changed to create the two types of classifiers. To do so, if current environmental state is in the aliasing state list $asl$, and the memory list $ml$ is not empty, then the covering procedure will create classifiers with memory condition part; otherwise some classifiers without memory conditions will be created in the same way as the original XCS, and their memory pointer $mp$ will be set to '-1', which indicates they are classifiers without memory conditions. The method of creating a classifier $cl$ with memory condition is as follows.

First, the normal condition part of this classifier $cl$ will be created using the current environmental state in the same way as the original XCS.

Second, if the first element $ml[0]$ of the memory list is not an aliasing state, then the memory condition part of $cl$ will be created using $ml[0]$, and the memory pointer $mp$ of $cl$ will be set to '0'. Otherwise, the second element $ml[1]$ will be considered in the same way. If $ml[1]$ is also an aliasing state, the third element $ml[2]$ will be considered. And so forth, until an element that is not an aliasing state is found, or all elements of the memory list have been checked and they are all aliasing states. If every element is an aliasing state, we still need to select an element to cover the memory condition part of $cl$. To do so, an element is randomly selected from the aliasing state list with probability inverse proportional to its parameter $num$; or the element with the smallest $num$ is selected deterministically. If the element $ml[i]$ is found or selected from the aliasing state list, the memory condition part of $cl$ will be created using $ml[i]$ in the same way as the original XCS, and the memory pointer $mp$ of $cl$ will be set to the value of $i$, to indicate which element in the memory list the memory condition of $cl$ will match.

Next, the description of matching process will be given. The matching procedure in XCS is changed a little to adapt it to the architecture of XCSMD. If the current environmental state is an aliasing state, and the memory list $ml$ is not empty, then classifiers with memory conditions will be matched by the current environmental state and the memory list together, to form the match set [M]; otherwise, classifiers without memory conditions will be matched by the current environmental state to form the match set [M] in the same way as the original XCS. It should be noted that, which element in the memory list will be matched by the memory condition of classifier $cl$ is determined by the memory pointer $mp$ of $cl$.

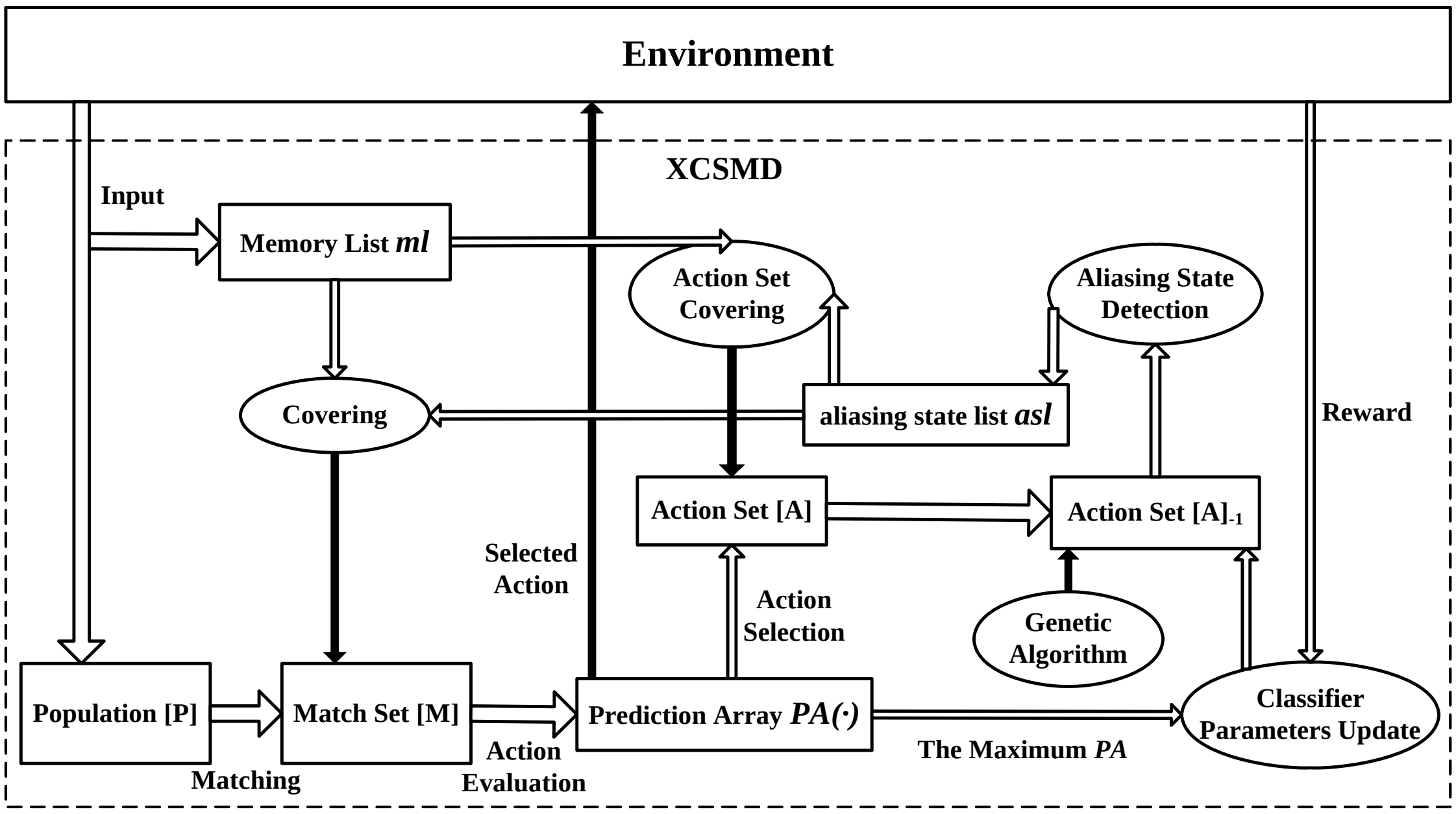

**Fig. 5** The main process of XCSMD.

## 4.5 The Genetic Algorithm in XCSMD

The genetic algorithm procedure in XCSMD is similar to that in the original XCS with only some little differences. First, there are two types of classifiers in XCSMD, thus they need to be treated separately in some respects, since one type is prepared for aliasing states while the other type is mainly for non-aliasing states in the environment. When GA selects the parents to reproduce, two classifiers of the same type are selected from action set [A], and the offspring is created out of them. If there is only one classifier in action set, the parents will be the same as this classifier. When the offspring classifier includes the memory condition, crossover can occur at any point in the string formed by concatenating the characters representing the normal condition and memory condition. Mutation, same as crossover, can also occur at any point in the string. Additionally, mutation can also apply to the memory pointers of classifiers with memory conditions, to make the value of the memory pointer $mp$ increase (+1) or decrease (-1). Finally, if GA subsumption is being used, subsuming classifier and subsumed classifier not only must be the same type of classifiers, but also the values of their memory pointer $mp$ should be equal. This is also the case with action set subsumption.

## 4.6 Summary about the Process of XCSMD

XCSMD works basically like XCS, although there are two types of classifiers in it. At every time step, match set [M], system prediction array $PA(\cdot)$, and action set [A] are formed essentially as in XCS. The differences are that memory conditions must also be matched if necessary when building [M] in XCSMD. The classifier's parameters are updated as in XCS. At the beginning of each experiment, the aliasing state list is set to empty. The aliasing state detection procedure is responsible for the formation of the aliasing state list, while the action set covering operator provides enough aliasing state relevant candidate classifiers for the detection mechanism. Fig. 5 shows the main process of XCSMD. At the beginning of each problem (trial), the internal memory list is initialized setting to empty. The update of the memory list is simple, as long as we take it as a queue data structure, which follows the first-in-first-out rule. The genetic algorithm procedure in XCSMD should treat the two types of classifiers equally and separately in some respects. About the other processes, there is no difference between XCSMD and XCS.

In XCSMD, the aliasing state detection and disposal are integrated as a whole, which is different from the parallel and distributed approach of Hamzeh's Parallel Specialized XCS (PSXCS) and Recursive PSXCS (RPSXCS). There are some special sub-XCSs in PSXCS. One of the sub-XCSs is to detect aliasing states in the environment and responsible for finding optimal actions for non-aliasing states. The other sub-XCSs equipped with history windows are responsible for each detected aliasing state respectively. Besides, a control unit is used to coordinate these parallel and distributed sub-XCSs in PSXCS. The approach of RPSXCS is similar to that of PSXCS. Therefore, it can be seen that the structure of XCSMD is simple and compact compared with PSXCS (RPSXCS), though all of them have employed aliasing state detection and history window.

The history window approach is a resource-consuming method since the use of the history window makes the condition of classifiers become longer and the search space becomes larger. But our proposed method, like the approach of PSXCS (RPSXCS), limits the scope of the history window mainly to aliasing parts of the search space, since the classifiers with memory conditions are responsible for aliasing states while the classifiers without memory conditions are mainly for non-aliasing states. In XCSMD, the history window size of classifiers with memory conditions is just '1', and classifiers without memory conditions have no history window. So the window size is further reduced. This is different from that of PSXCS (RPSXCS), where the history window size is always bigger than '1'. Besides, aliasing state detection method employed by XCSMD is partially inherited from Dung et al.(Dung et al. 2008), which is also different from the method adopted by PSXCS (RPSXCS) in many respects. In Section 6, we will see that the approach of XCSMD to handle partially observable environment is effective and efficient.

In addition, XCSMD's memory list element and memory condition is longer than the memory register of XCSMH. However, experiments below will show that, it does not bring obvious draw-backs, because the generalization mechanism, which is inherited from XCS, can quickly get rid of those irrelevant attribute values in classifiers' memory conditions.

# 5. Experimental Setup

## 5.1 Policy of Experiments

In the maze problems, the experiment typically consists of a number of problems (trials) that the agent must solve, and for each problem, the agent is placed into a randomly chosen empty square. Then the agent moves under the control of the classifier system avoiding obstacles until either it reaches the food or had taken $M_{es}$ steps, at which point the problem ended unconditionally as done in Lanzi(Lanzi 1999). However, in this paper, $M_{es}$ is only a parameter to control the maximum number of permissible time steps, which is used to avoid the agent falling into a loop. The agent will not change its position if it chooses an action to move to a square with the obstacle inside, though one time-step still elapses.

We will employ the typical exploration/exploitation strategy used with XCS by Wilson(Wilson 1995), Lanzi(Lanzi 1998a; Lanzi and Wilson 2000) and Butz(Butz and Wilson 2002). To deal with these trials, the agent will solve them in exploration or in exploitation approximately alternately, that is learning and testing approximately alternate. When in exploration problems, the system decides, with a probability $P_s$ (a typical value is 0.5), whether to select actions randomly or to choose the action that predicts the highest payoff. However, when in exploitation problems, the genetic algorithm is turned off and the action which predicts the highest payoff is always selected. System performance is computed as the average number of steps to food in the last 50 exploitation problems. And in order to evaluate the final policy evolved, in each experiment, exploration is turned off during the final 2500 problems and the system works only in exploitation. This is same as Lanzi's methods(Lanzi 1998a; Lanzi and Wilson 2000). Unless otherwise specified, every statistic results presented in this paper is averaged on ten experiments.

## 5.2 Setting of Learning Classifier System

The following classifier structure was used for LCS in the experiments: Each classifier has 16 binary bits in the normal condition field and memory condition field respectively: two bits for each of the 8 neighboring squares, with 00 representing the situation that the square is empty, 11 that it contains food (F), and 01 that it is an obstacle (T). Also similar to Wilson's(Wilson 1995) and Lanzi's(Lanzi 1998a; Lanzi and Wilson 2000) experiment setting, the following LCS's parameter values are used for both XCSMD and XCS, unless otherwise specified: $\alpha$=0.1, $\beta$=0.2, $\gamma$=0.71, $\varepsilon_0$=5.0, $\nu$=5.0, $\theta_{GA}$=25, $\chi$=0.8, $\mu$=0.01, $\delta$=0.1, $\theta_{del}$=25, $\theta_{sub}$=35, $\theta_{mna}$=8, $p_I$=10, $F_I$=0.01, $\varepsilon_I$=0, $P_s$=0.5, $P_{\#}$=0.3, $M_{es}$=100, $\theta_{ASR}$=30, $\tau$=0.4, $\theta_{ASCover}$=20, $N_{ml}$=5. The size of population of the classifiers will be specified in each experiment below.

The whole system proposed by us is developed based on Butz's XCS open source code(Butz 2003), which can be downloaded from the Internet.

# 6. Experiments and Analysis in Maze Environments

Four types of different complex maze problems are tested and studied here, to illustrate the generality of our approach. We will follow Zatuchna's metric(Bagnall and Zatuchna 2005) for classifying the types of aliasing, which have different influence on maze complexity, and thus for classifying the types of maze environment. The first test case is Woods1, a typical simple Markov maze environment; and then two typical Type-I aliasing mazes will be tested; the third is two Type-II aliasing mazes; the last is four Type-III aliasing mazes, which may be considered as the most difficult aliasing mazes.

## 6.1 Experiments in Markov Environment

In the first experiment, we applied XCSMD to the Woods1 environment (Fig. 1). Woods1 is a typical Markov environment for testing the learning system, since no aliasing state exists in the environment. In Woods1, the optimal average path to the food is 1.6875 steps. This experiment is used to show that XCSMD can solve the general maze problem.

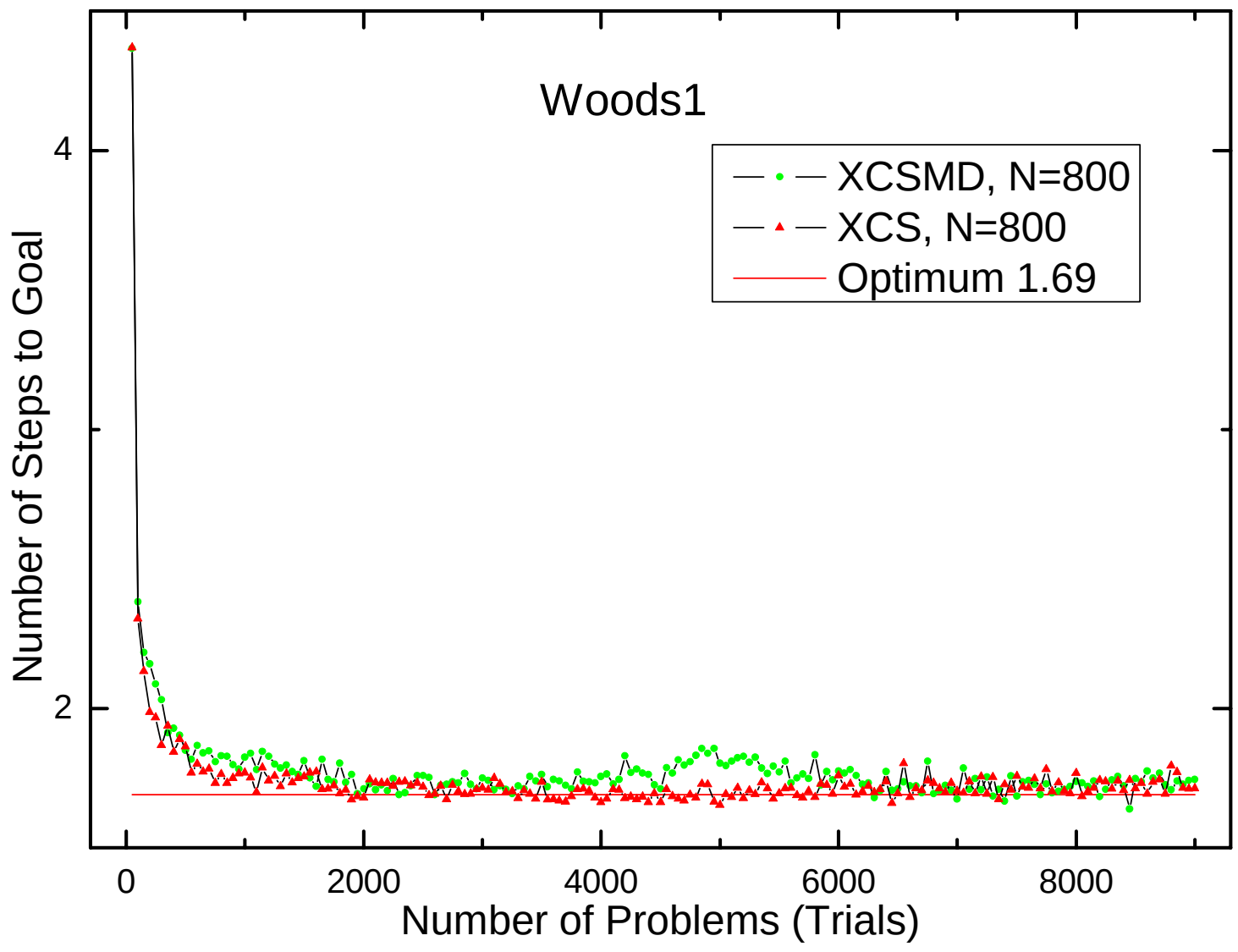

**Fig. 6** The results of applying XCSMD to Woods1, compared with XCS. Population size is set to 800 classifiers. The curve is averaged over ten experiments. The optimum is 1.69 steps.

XCSMD used a population size $N$ of 800 classifiers, while XCS used the same population size for comparison. Fig. 6 shows the comparison between the performance of XCS and XCSMD in Woods1. In both cases, the results converge to the optimal solution, about 1.7. XCSMD can solve the general maze problem as XCS. In addition, the experiment also shows that XCSMD converges slightly slower than XCS, since XCSMD is required to detect aliasing states. Thus a small number of non-aliasing states may be mistakenly identified as aliasing ones because of the influences of some stochastic processes in the system. Therefore, a few classifiers with memory conditions are created still in this case, which is the reason for the slightly slower convergence.

## 6.2 Experiments in Type-I Aliasing Mazes

This section, we will show the results of applying XCSMD to two typical Type-I aliasing mazes: MiyazakiA in Fig. 3(a) and Littman57 in Fig. 2. These experiments are used to test the XCSMD's performance in Type-I aliasing environments.

MiyazakiA maze, contains four aliasing states marked with 1, 2, 3, and 4, and each aliasing state with two aliasing squares, which are marked with "a" and "b" respectively. The optimal average path to the food is 3.05 steps. Both XCSMD and XCS used a population size $N$ of 2400 classifiers respectively.

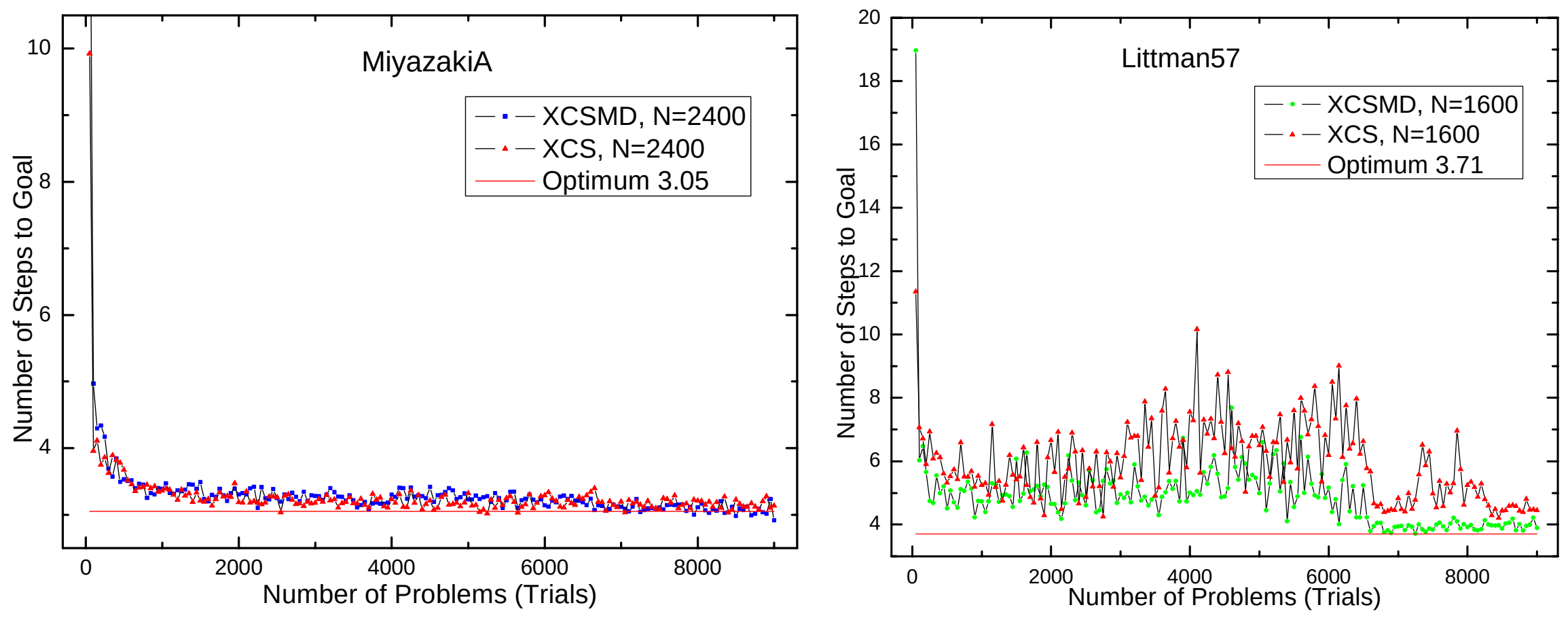

**(a)** The results in MiyazakiA **(b)** The results in Littman57

**Fig. 7** The results of applying XCSMD to MiyazakiA and Littman57, compared with XCS. The curve is averaged over ten experiments.

Fig. 7(a) shows the results for MiyazakiA. XCS's performance converges to an average of 3.16 steps to find food, evolving a near optimal solution. XCSMD with 2400 classifiers converges to an average of 3.22 steps to food within 6500 learning problems, and evolves an almost optimal solution during the last 2500 exploitation problems, about 3.08.

Fig. 7(b) shows the results for another Type-I aliasing maze, Littman57. It contains three aliasing states marked with 1, 2, and 3, and each aliasing state with three aliasing squares, which are marked with a, b, and c respectively. Population size is set to 1600 classifiers for both XCSMD and XCS. The optimal average path to the food is 3.71 steps. The average number of steps for AgentP to find food in Littman57 is 4.82 steps(Zatuchna and Bagnall 2009). For this case study, the classifier sets found by XCS are suboptimal, about 4.94, and the result of XCSMD converges to a near-optimal solution, around 3.95 steps. Note that, during the learning period (first 6500 problems), the performances oscil-late heavily above the optimum, although the fact that the curve records performance on test problems only; during the test period (last 2500 problems), the oscillation still exists but slightly.

The reason that XCSMD as well as XCS performs less well and have some fluctuation in Littman57 seems because of the reinforcement learning mechanisms employed by XCS, and also by XCSMD. As shown in Formula (2), the Q-learning reinforcement procedure has been widely used in LCS(Stolzmann 2000; Wilson 1994, 1995). However, the way by which the Q-coefficients depend on the distance $d$ to food in a maze environment may lead to some disadvantages(Zatuchna 2005; Zatuchna and Bagnall 2009). In mazes, each action of the agent may result into three possible outcomes: the distance to food will shorten $(d-1)$, lengthen $(d+1)$, or remain unchanged. However, the influences of some stochastic processes in the agent or some aliasing states in environments may lead to the situation where the actual Q-learning coefficients may significantly fluctuate around their rated values(Zatuchna 2005; Zatuchna and Bagnall 2009). For example, if there is some probability that upon increasing $d$, the actual Q-learning coefficient for the optimal action may become smaller than the actual Q-learning coefficients for the non-optimal actions. Therefore, the agent may select the non-optimal actions in this case, rather than the optimal action. In our experiment with Littman57 maze, there are three aliasing squares for each aliasing state, i.e. the distance to the food is different in the three aliasing squares, but the same classifier is used at these squares, and the discounted reward as well as the Q-learning coefficients will be different for each application of this classifier. That is to say, XCS and XCSMD will use these different Q-learning coefficients to update the payoff prediction $p$ of the same classifier, as shown in Formula (3). Thus the prediction $p$ has not a good chance to converge, on the contrary, bounces between multiple different values. So, non-optimal actions may be taken as the optimal action mistakenly, which causes the fluctuation in the performance.

In addition, Littman57 has some aliasing conglomerates of a similar graphic pattern that include more than two aliasing states and located in different areas of the maze, i.e. aliasing clones, which is an even more complex phenomenon, and is a partial reason for the more difficulty with Littman57.

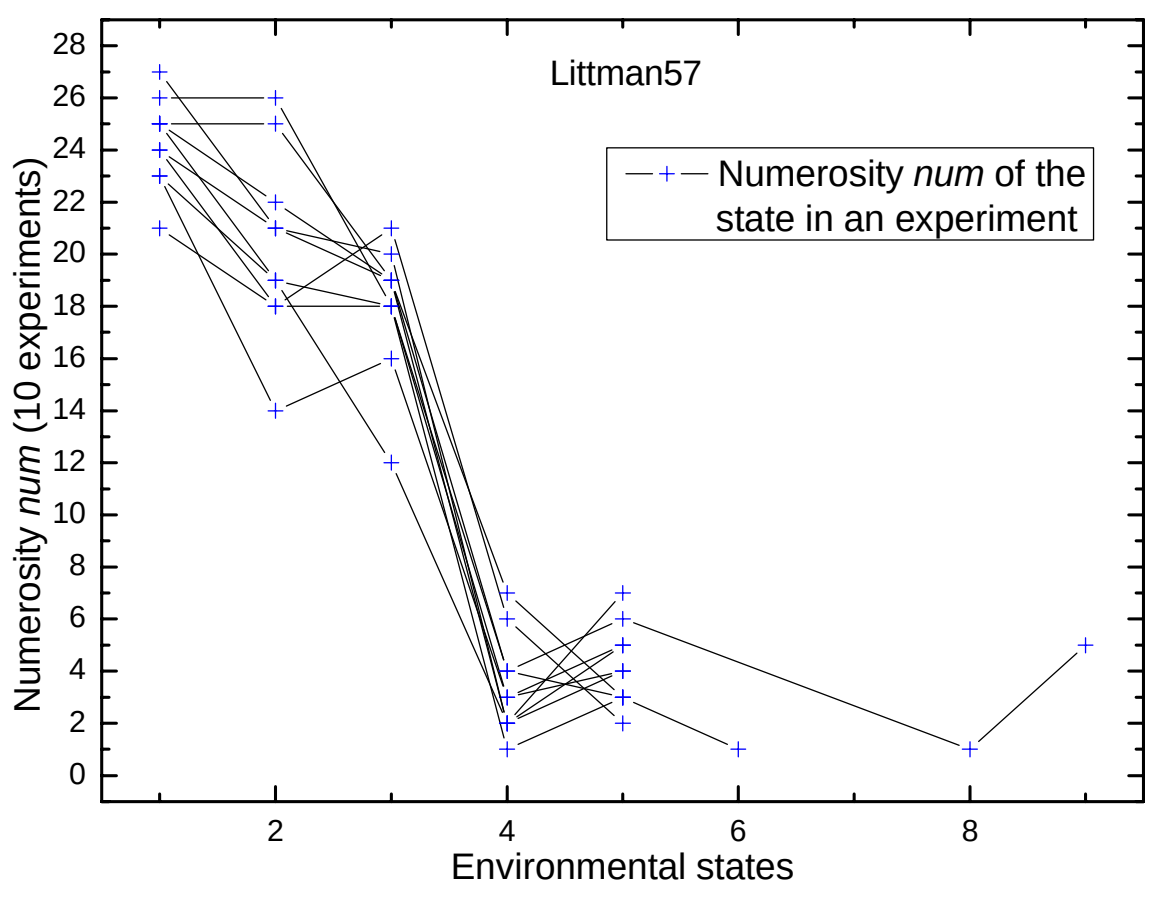

**(a)** Distribution

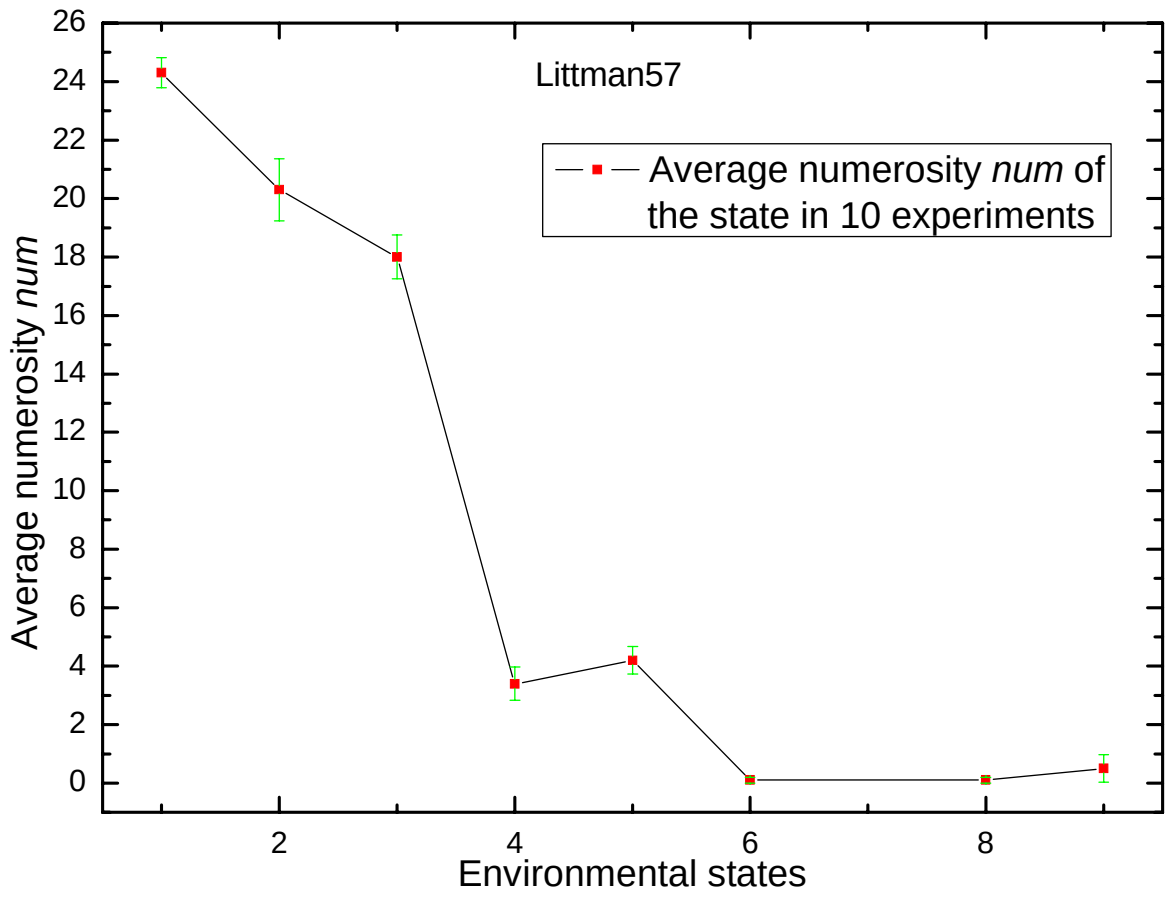

**(b)** Average and standard error

**Fig. 8** (a) The *numerosity* $num$ of each state over 10 experiments in Littman57. The *numerosity* indicates how many times the state has been recognized as an aliasing state in an experiment. The environmental state labels are shown in Fig. 2. (b) Average and standard error of the results shown in (a).

Figures 8 shows the performance of aliasing state detection in XCSMD for Littman57. Ten curves in Fig. 8(a) correspond to ten experiments. Each cross in the figure represents the number of times a state has been added to the aliasing state list $asl$ in an experiment. There are up to ten crosses for an environmental state, since some states may not be recognized as aliasing states in some experiments. The curve in Fig. 8(b) is the average and standard error result of the ten curves shown in Fig. 8(a). It should be noted that the aliasing state detection may mistakenly identify some non-aliasing states as aliasing ones because of the influences of some stochastic processes in the system. But the numerosity values of these non-aliasing states are significantly lower than the ones of aliasing states. Additionally, the standard error shown in Fig. 8(b) is small. Thus, the three aliasing states in Littman57 can be distinguished from the six non-aliasing states by their different numerosity values. So XCSMD can detect the three aliasing states correctly, which confirms the effectiveness of the aliasing state detection in XCSMD.

## 6.3 Experiments in Type-II Aliasing Mazes

Two typical Type-II aliasing maze environments will be studied and tested here, to illustrate the performance of our approach in Type-II aliasing mazes. The first is Maze7(Lanzi 1998a) in Fig. 4 (b), another is MazeF4 in Fig. 3(b). The two mazes with type II aliasing squares marked with 1a and 1b, which have different distances to food as well as different directions, are typical test instances for LCS. Besides, in the two mazes, the optimal solution requires the agent to visit more aliasing positions before it reaches the food, and performs longer sequences of actions to reach the goal state. So, they are rather complex to some LCSs.

The optimal average path to the food is 4.33 steps in Maze7, and 4.50 in MazeF4. XCSMD and XCS both used a population size $N$ of 1600 classifiers in Maze7 and in MazeF4. In the two experiments, the maximum number of permissible time steps $M_{es}$ of one trial in XCS was set to 20.

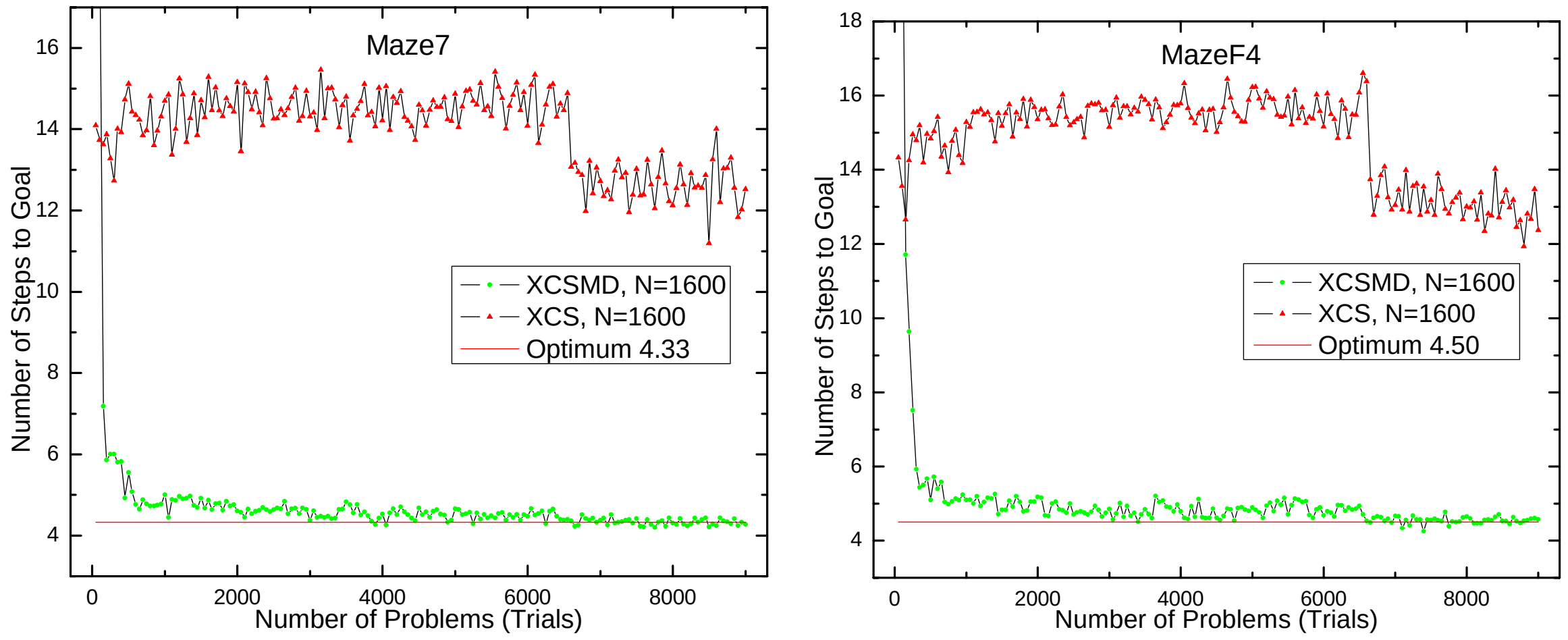

**(a)** The results in Maze7 **(b)** The results in MazeF4

**Fig. 9** The results of applying XCSMD to Maze7 and MazeF4, compared with XCS. The curve is averaged over ten experiments.

Fig. 9(a) shows the comparison between the results of XCS and XCSMD in Maze7 environment, and Fig. 9(b) in MazeF4 environment. In both cases, the results of XCSMD converge to the almost optimal solution within 6500 learning problems, and evolve the optimal solution during the last 2500 exploitation problems, while XCS completely fails. The classifier system AgentP in Maze7 and MazeF4 can also get optimum(Zatuchna and Bagnall 2009). From other literature, we can only know that, Lanzi's XCSMH1 can get the optimal solution(Lanzi and Wilson 2000) in Maze7 under similar conditions as our XCSMD here, and Stolzmann's ACS can get the optimal solution(Stolzmann 2000) in MazeF4.

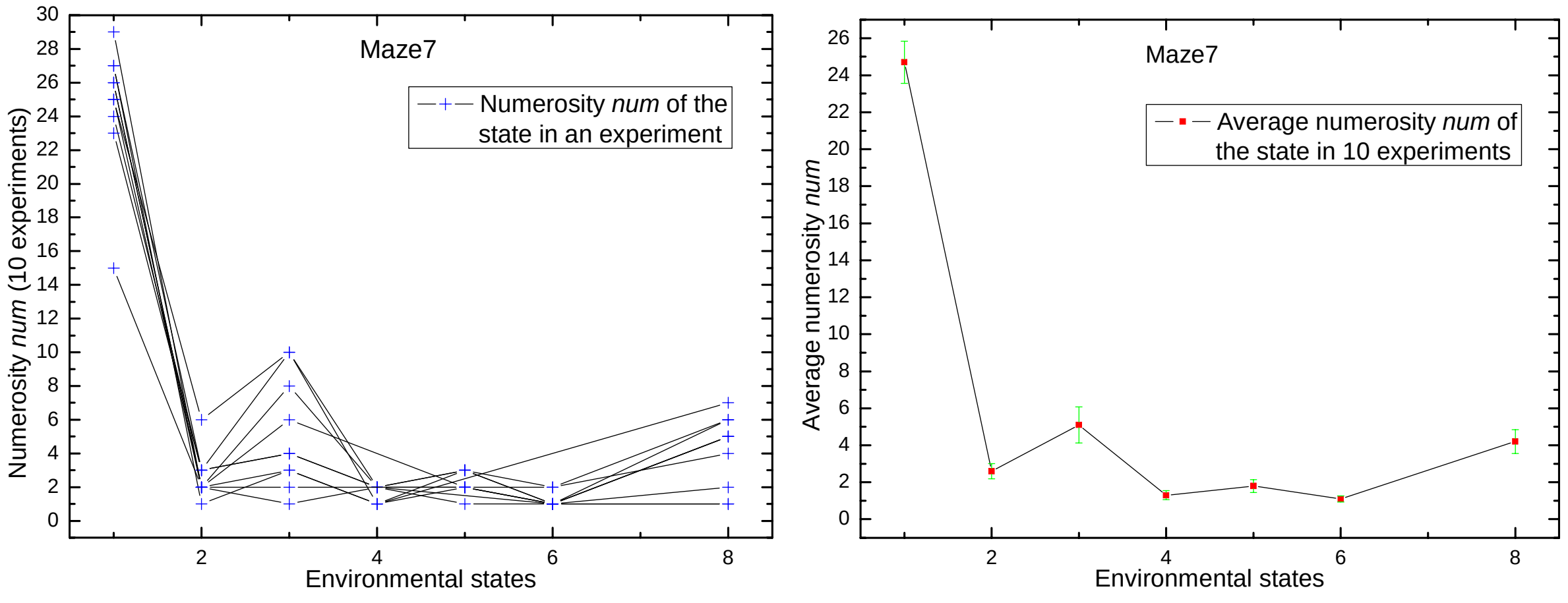

**(a)** Distribution **(b)** Average and standard error

**Fig. 10** (a) The *numerosity* $num$ of each state over 10 experiments in Maze7. The *numerosity* indicates how many times the state has been recognized as an aliasing state in an experiment. The environmental state labels are shown in Fig. 4. (b) Average and standard error of the results shown in (a).

Figures 10 (a) and (b) show the results of aliasing state detection in XCSMD for Maze7. The numerosity values of these non-aliasing states are significantly lower than that of the only one aliasing state. And the standard error with these numerosity values is small. XCSMD can recognize the aliasing state in Maze7 effectively. This is also the case with MazeF4, and the result about MazeF4 will not be presented here due to space constraints.

About the two mazes, XCSMD's memory mechanism can transform them into Markov environments almost completely, so XCSMD can solve them. Generally, Type-II Aliasing maze environments are simple for XCSMD.

## 6.4 Experiments in Type-III Aliasing Mazes

In the last set of experiments, we will apply XCSMD to four Type-III aliasing mazes: Woods101 in Fig. 1(b), Woods101$\frac{1}{2}$(Lanzi and Wilson 2000) in Fig. 11(a), Woods102 in Fig. 4(a), and Maze10(Lanzi 1998b) in Fig. 11(b). First, the experimental results with the first three mazes will be described, and then the results about Maze10 will be presented later.

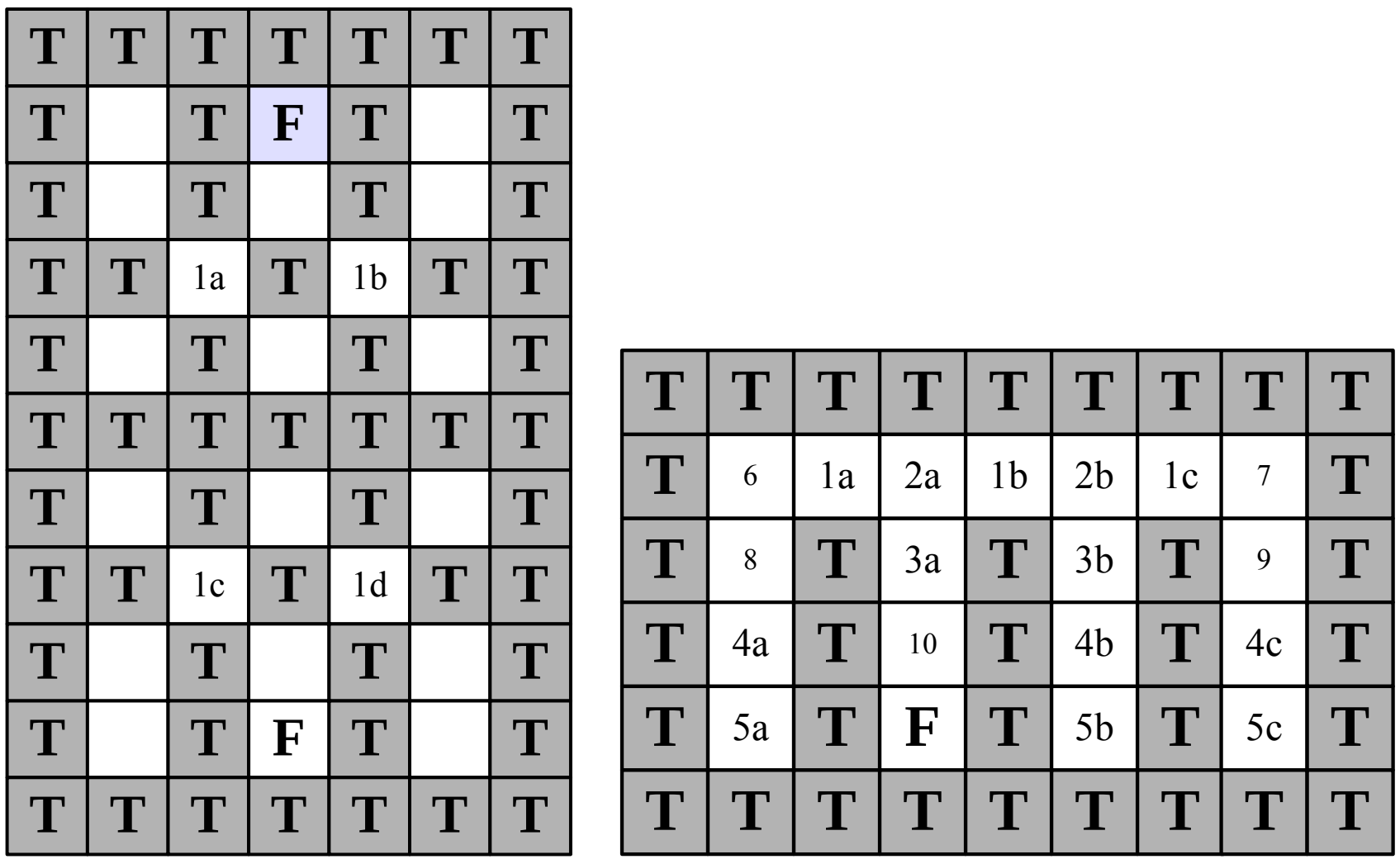

**(a)** **(b)**

**Fig. 11** (a) Woods101$\frac{1}{2}$ environment; (b) Maze10 environment.

The Woods101 environment has two aliasing squares which return the same sensory configuration and have same distance to the food, but require different optimal actions. The Woods101$\frac{1}{2}$ environment has four such aliasing squares belonging to a same aliasing state, and Woods102 also has four such squares belonging to an aliasing state, and two other squares belonging to another aliasing state. The optimal average path to the food in Woods101 is 2.90 steps, 3.10 steps in Woods101$\frac{1}{2}$, and 3.308 steps in Woods102. The parameter $N$ of population size on the three mazes will be shown in the performance graphs. In the three experiments, the maximum number of permissible time steps $M_{es}$ of one trial in XCS was set to 20.

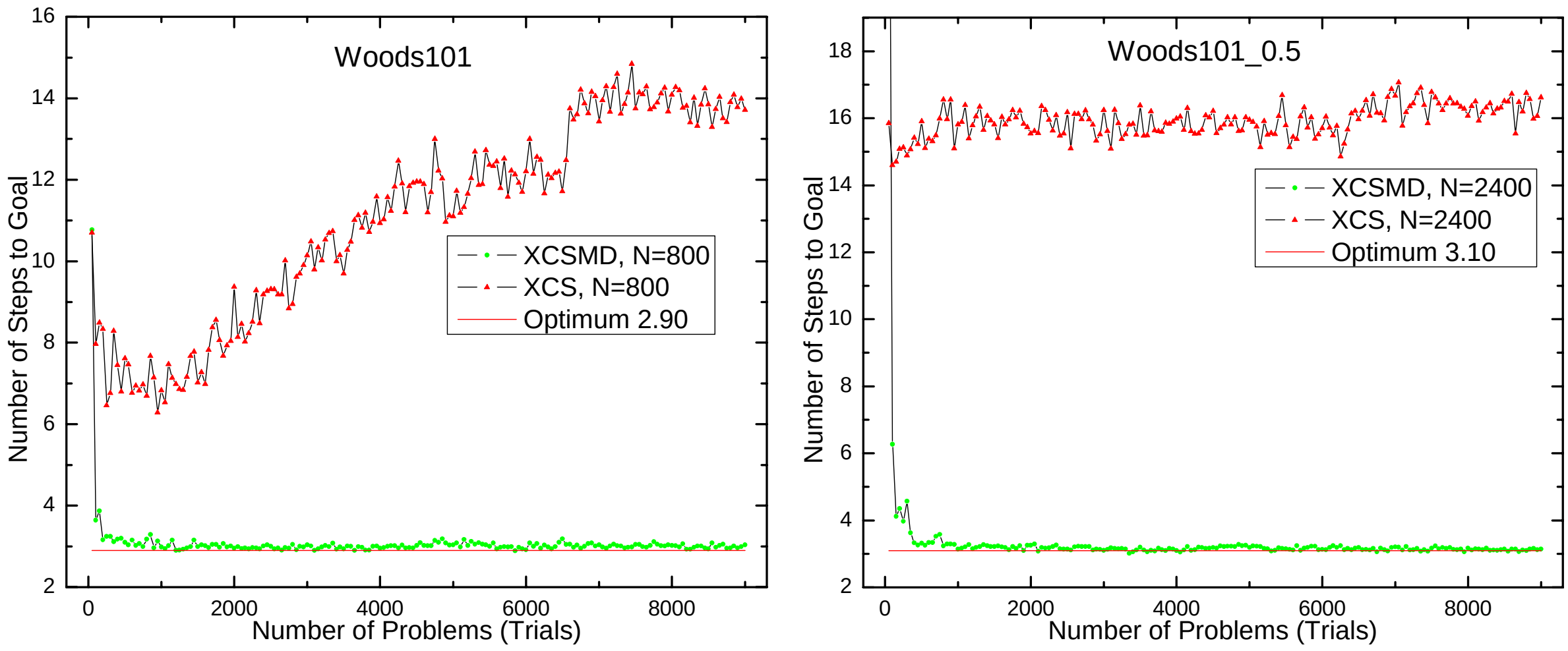

**(a)** The results in Woods101 **(b)** The results in Woods101$\frac{1}{2}$

**Fig. 12** The results of applying XCSMD to Woods101 and Woods101$\frac{1}{2}$, compared with XCS. The curve is averaged over ten experiments.

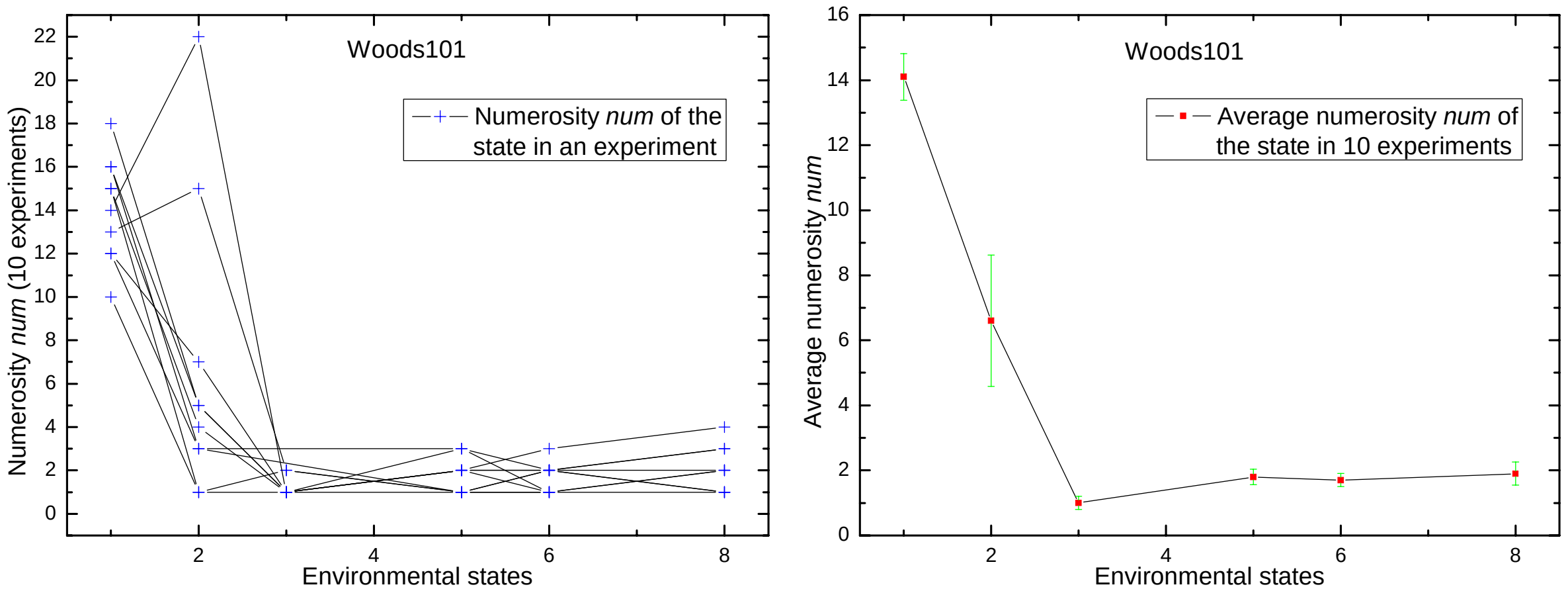

**(a)** Distribution **(b)** Average and standard error

**Fig. 13** (a) The *numerosity* *num* of each state over 10 experiments in Woods101. The *numerosity* indicates how many times the state has been recognized as an aliasing state in an experiment. The environmental state labels are shown in Fig. 1. (b) Average and standard error of the results shown in (a).

As can be seen from Fig. 12(a), Fig. 12(b), and Fig. 14(a), in the three maze environments, corresponding XCSMD's results converge to near optimal solutions within 6500 learning problems, and evolve almost optimal solutions during the last 2500 exploitation problems, while XCS can do nothing to them. From Lanzi's literature, XCSM1 with 1600 classifiers(Lanzi 1998b) or XCSMH3 with 800 classifiers(Lanzi and Wilson 2000) can get the optimal solution in Woods101, and XCSMH4 with 2800 classifiers can get an almost optimal solution(Lanzi and Wilson 2000) in Woods101$\frac{1}{2}$ (about 3.3 steps), while in Woods102, XCSMH8 with 6000 classifiers can get optimum(Lanzi and Wilson 2000) within 35000 learning problems, and XCSM2 with 2000 classifiers can get near optimum, about 4.1 steps(Lanzi 1998b). The classifier system AgentP in the three maze environments can also get optimum(Zatuchna and Bagnall 2009).

Figures 13 shows the results of aliasing state detection in XCSMD for Woods101. In ten experiments, the numerosity value of the only one aliasing state is usually bigger than the ones of non-aliasing states. However, from Figures 13(a), the numerosity value of the environmental state 2 is sometimes bigger than that of the aliasing state 1. Note that state 2 is actually a pseudo-aliasing state, which includes two pseudo-aliasing squares with the same distance and action to the goal. Because of the influences of some stochastic processes in XCSMD, the aliasing state detection may mistakenly identify the pseudo-aliasing state as aliasing one. However, this has no effect on the performance of XCSMD in Woods101, since only a pseudo-aliasing state is identified as an aliasing one, rather than vice versa. So the aliasing state in Woods101 can be recognized effectively by XCSMD. This is also the case with Woods101$\frac{1}{2}$ and Wood102, and their results will not be presented here.

Similar to Type-II aliasing maze environments, the three mazes can also be transformed into Markov environments by XCSMD's memory mechanism. XCSMD is capable of solving them.

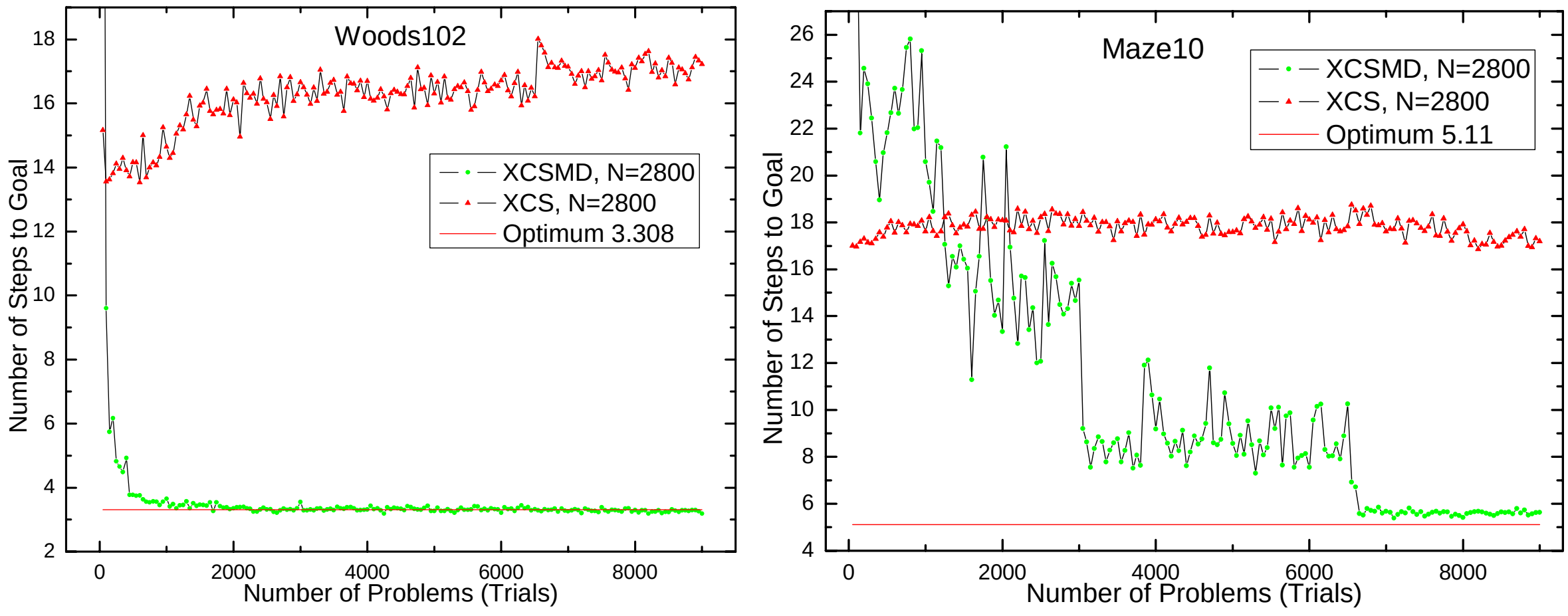

**(a)** The results in Woods102

**(b)** The results in Maze10

**Fig. 14** The results of applying XCSMD to Woods102 and Maze10, compared with XCS. The curve is averaged over ten experiments.

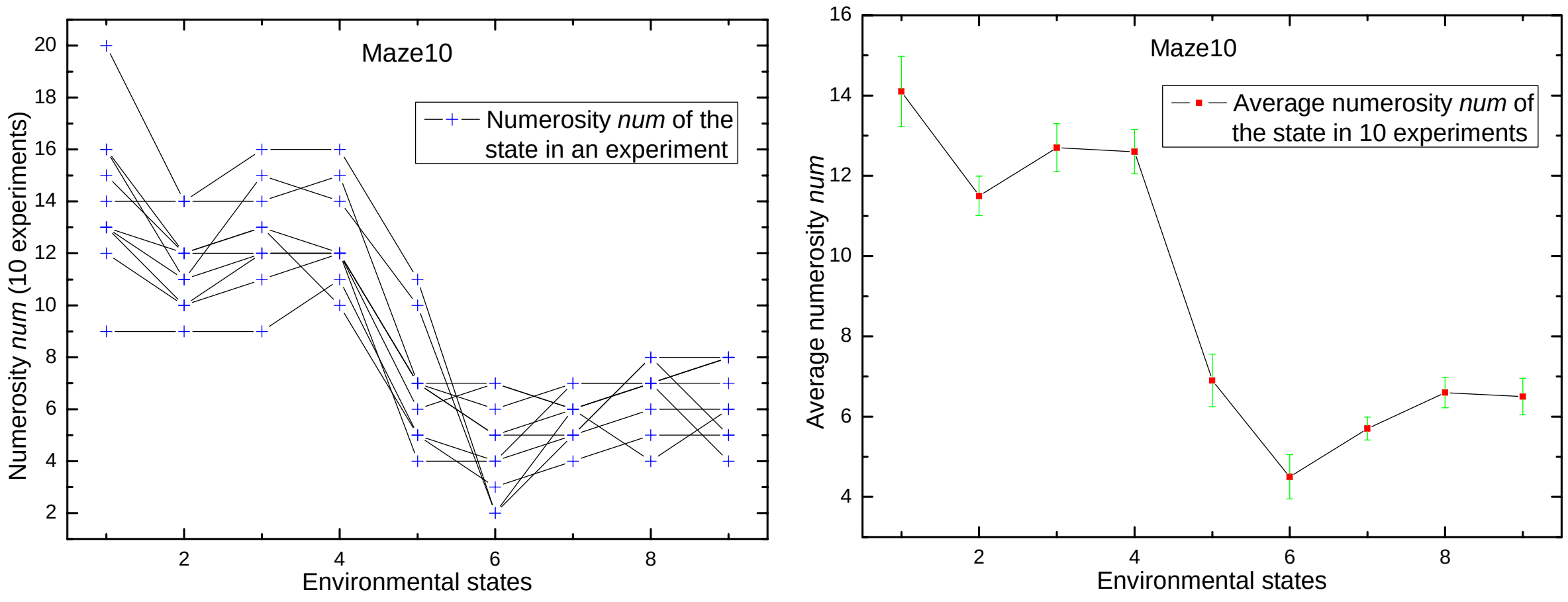

**(a)** Distribution

**(b)** Average and standard error

**Fig. 15** (a) The *numerosity* $num$ of each state over 10 experiments in Maze10. The *numerosity* indicates how many times the state has been recognized as an aliasing state in an experiment. The environmental state labels are shown in Fig. 11(b). (b) Average and standard error of the results shown in (a).

Next, we will show the experiment about Maze10. The population size on this maze was set to 2800. The optimal average path to the food in Maze10 is 5.11 steps. The parameter $M_{es}$ in XCS was set to 20.

As is shown in Fig. 14(b), XCSMD can evolve a near optimal solution (about 5.60 steps) in Maze10 while XCS failed. Lanzi's XCSMH1 with 2000 classifiers can also get a near optimal solution(Lanzi 1998b) (about 6.4 steps) under similar conditions here, and Zatuchna's AgentP can get suboptimum(Zatuchna and Bagnall 2009), about 7.87 steps.

In fact, Maze10 is a quite difficult environment, and more complex than other mazes presented in this paper. As it can be noticed, Maze10 is a variant of Woods101, and similar to Maze7, needs longer sequences of actions to reach the goal state, and it has four types of aliasing squares: 1b and 1c, 4a and 4c, 4b and 4c, 5a and 5c, 5b and 5c are type I aliasing squares; 2a and 2b, 1a and 1c, 3a and 3b are type II aliasing squares; 1a and 1b are type III aliasing squares; 4a and 4b, 5a and 5b are pseudo-aliasing squares.

Besides, there are two aliasing clones involving 1-2-3 and two ones involving 3-2-1 states in Maze10. Therefore, the optimal solution requires the agent to visit more aliasing squares, and pass through some aliasing clones before it reaches the food. From the mechanism of XCSMD, we can know that the agent can pass through the aliasing clones in Maze10, because the classifier's memory condition of XCSMD as a foothold outside the aliasing clones can determine the exact location of the agent, and also can decide which direction the agent should go along, since the memory condition can match the earlier environmental state, not just the state at the last time step. In addition, aliasing clones also exist in Littman57 and Woods102 environments, but their cases are different. In each aliasing clone of Littman57, the direction that the agent should go along is the same, despite the agent may not determine its exact location. So the agent just needs to go along this direction, until passing through the aliasing clones. In Woods102, the two aliasing clones are two chains of adjacent aliasing squares. The optimal solution just requires the agent to visit one aliasing square, that is the agent does not need to pass through the whole aliasing clones, just needs to skim them. However in Maze10, the agent controlled by XCSMD is impossible to keep away from the aliasing clones, and has to pass through them. So Maze10 is more difficult than Littman57 and Woods102, which is the main reason for XCSMD only evolving near optimal solution in it.

Figures 15 (a) and (b) show the results of aliasing state detection in Maze10. From Figures 15(a), the numerosity values of the first four aliasing states are significantly bigger than the ones of non-aliasing states. However, the numerosity value of the aliasing state 5 is not always bigger than the ones of non-aliasing states. So, the state 5 cannot be consistently distinguished from the non-aliasing states by its numerosity value, and XCSMD may identify it as a non-aliasing state sometimes. In fact, state 5 is a special state in Maze10 since it is farthest from the goal state and belongs to both type I aliasing state and pseudo-aliasing state, which causes great difficulties to distinguish it. This also reveals the difficulty with Maze10 on the other hand, and seems to be a partial reason for XCSMD's non-optimal performance in Maze10.

## 6.5 Summary about the Experiments

Following Zatuchna's metric(Bagnall and Zatuchna 2005) for classifying the types of aliasing problems and thus classifying the types of maze environments, four types of different complex maze problems have been tested by our approach and XCS, involving nine mazes in total. Here, we will present a summary of the comparison between our system and some other LCSs on these nine mazes, shown in Table 1.

**Table 1** Performance comparison between XCSMD and some other LCSs on the nine mazes

| Mazes\LCSs | Aliasing Type | Optimum | XCS (N) | XCSM, XCSMH (N) | RPSXCS (N) | XCSMD (N) |
|---|---|---|---|---|---|---|
| Woods1 | non-aliasing | 1.69 | 1.72 (800) | XCSMH0, 1.68(800)(Hamzeh et al. 2009) | 1.68(800)(Hamzeh et al. 2009) | 1.72 (800) |
| MiyazakiA | I | 3.05 | 3.16 (2400) | — | — | 3.08 (2400) |
| Littman57 | I | 3.71 | 4.94(1600) | — | — | 3.95 (1600) |
| Maze7 | II | 4.33 | failed (1600) | XCSMH1, 4.3 (1600)(Lanzi and Wilson 2000) | 4.34(>1500)(Hamzeh et al. 2009) | 4.33 (1600) |
| MazeF4 | II | 4.50 | failed (1600) | XCSMH1, 4.59 (2000)(Hamzeh et al. 2009) | 4.50(>2000)(Hamzeh et al. 2009) | 4.55 (1600) |
| Woods101 | III | 2.90 | failed (800) | XCSM1, 3.0 (1600)(Lanzi 1998b); XCSMH3, 2.9 (800)(Lanzi and Wil-son 2000) | — | 3.00 (800) |
| Woods101 $\frac{1}{2}$ | III | 3.10 | failed (2400) | XCSMH4, 3.3 (2800)(Lanzi and Wilson 2000) | 3.11(>4000)(Hamzeh et al. 2009) | 3.14 (2400) |
| Woods102 | III | 3.31 | failed (2800) | XCSM2, 4.1 (2000)(Lanzi 1998b); XCSMH8, 3.23 (6000)(Lanzi and Wilson 2000) | 3.31(>4000)(Hamzeh et al. 2009) | 3.28 (2800) |
| Maze10 | III | 5.11 | failed (2800) | XCSMH1, 6.4 (2000)(Lanzi 1998b) | — | 5.60 (2800) |

The table of performance comparison represents the best available statistics on the nine mazes to our knowledge for learning agents in these LCS groups. Population size $N$ is shown in the parenthesis. The number behind "XCSM" or "XCSMH" is used to indicate the size (in bits) of internal memory employed by XCSM or XCSMH. In "RPSXCS" column of the table, ">" indicates the population size is greater than the given value, because RPSXCS contains some subsystems, and the given value is just the population size of one main subsystem. Some statistics in the table are approximate results obtained from the graphs of the original papers, since the precise average step-to-goal values are not always available from those papers.

# 7. Conclusions and Implications

The XCS classifier system has solved the former main shortcoming of LCSs, by its accuracy based fitness approach. However, XCS can only learn optimal policy in Markov environments. There-fore, a memory mechanism is required for XCS to cope with non-Markov problems, since most environments are non-Markov in practice.

Unlike Wilson's proposal and Lanzi's implement of XCSMH, XCSMD proposed by us adds an internal message list to XCS as the internal memory list, in which the length of a memory list element is the same length as strings from the detectors, and then only extend classifier with a memory condition, which is used to sense a specified element in the memory list. So, XCSMD holds a simple and clear memory mechanism. However, the overhead of applying memory conditions to the entire search space is a waste. Thus, XCSMD introduces two types of classifiers to reduce the space and time complexity of the system. One type is the normal classifier which is the same as that in XCS, to cope with non-aliasing states. The other type consists of memory condition part, normal condition part and action part, and is prepared to deal with aliasing states. So, in XCSMD, most non-aliasing states in problems are still handled as in XCS, which is an efficient way to reduce the complexity of the search space. Besides, XCSMD resorts to an aliasing state detection method to recognize aliasing states in environments, which makes the memory mechanism of XCSMD become more efficient and effective.

In section 6, four sets of different complex maze problems have been tested by XCSMD. Results show that the aliasing state detection method employed by XCSMD is effective to recognize those aliasing states in environments; XCSMD can evolve optimal or near-optimal solutions in these difficult non-Markov environments, and outperform some other LCSs in some respects. Thus, XCSMD is a promising classifier system to solve non-Markov problems.

In performance, XCSMD is similar to RPSXCS, but needs much fewer classifiers than RPSXCS, especially in Woods102 and Woods101$\frac{1}{2}$. Seen from Table 1, XCSMD's performance is slightly better than XCSMH8 in Woods102, because XCSMH needs twice classifiers than XCSMD to evolve the optimal solution. In Maze10, XCSMD outperforms XCSMH, although they just get near optimum. XCSMD's memory condition is longer than the classifier's internal condition in XCSMH, but there is no significant difference in population size and number of learning trials required to evolve optimal solutions between them. The long memory condition does not bring obvious shortcomings, because the generalization mechanism of XCSMD can quickly get rid of those irrelevant attribute values in memory conditions of classifiers. And there are two types of classifiers in XCSMD, which further reduces the system overhead.

Additionally, some maze environments have aliasing clones, which make them become more complex. XCSMD can pass through these aliasing clones, because the classifier's memory condition of XCSMD as a foothold outside the aliasing clones can determine the exact location of the agent and which direction the agent should go along, since the memory condition can match the earlier environmental state, not just the state at the last time step. But unfortunately, some mazes have very complicated aliasing clones, which is an even more complex phenomenon, and is the main reason for XCSMD only evolving near optimal solution in Maze10. So the memory mechanism of XCSMD needs further improvement.

In our future work, we are going to examine XCSMD in some more complex and large maze problems, to improve its memory mechanisms further. Finally, one of the promising areas of future research seems to combine XCSMD with an “abstraction” algorithm and a “hypothesizing” algorithm proposed by Browne & Scott(Browne and Scott 2005). Abstraction attempts to find patterns in the classifiers that performed best within an LCS agent. And in reverse order, if the system finds a pattern common to two or more classifiers, the hypothesizing algorithm will generate some new classifiers using this pattern as a template. Therefore, this algorithm would allow the agent to extrapolate the knowledge obtained in one area of the environment to other areas of the same environment, or transfer the knowledge obtained in a certain environment to other similar environments(Iqbal et al. 2012; Zatuchna 2005). Thus the whole system will perform effectively and robustly.